\documentclass{nle}

\usepackage{amsmath}
\usepackage{graphicx}
\usepackage{natbib}
\usepackage{adjustbox}
\usepackage{float}
\ifpdf%
\usepackage{epstopdf}%
\else%
\fi
\usepackage{multirow}

\begin{document}
\label{firstpage}

\lefttitle{\LaTeX\ Supplement}
\righttitle{Natural Language Engineering}

\papertitle{Article}

\jnlPage{1}{00}
\jnlDoiYr{2019}
\doival{10.1017/xxxxx}

\title {Exploring Prompt-Based Methods for Zero-Shot Hypernym Prediction with Large Language Models}

\begin{authgrp}
\author{Mikhail Tikhomirov}
\affiliation{Lomonosov Moscow State University, Moscow, Russia\\
        \email{tikhomoriv.mm@gmail.com}\\}
\author{ Natalia Loukachevitch}
\affiliation{Lomonosov Moscow State University, Moscow, Russia\\ ISP RAS Research Center for Trusted Artificial Intelligence, Moscow, Russia \\
        \email{louk\_nat@mail.ru}}
\end{authgrp}

\history{(Received xx xxx xxx; revised xx xxx xxx; accepted xx xxx xxx)}

\begin{abstract}
This article investigates a zero-shot approach to hypernymy prediction using large language models (LLMs). The study employs a method based on text probability calculation, applying it to various generated prompts. The experiments demonstrate a strong correlation between the effectiveness of language model prompts and classic patterns, indicating that preliminary prompt selection can be carried out using smaller models before moving to larger ones. We also explore prompts for predicting co-hyponyms and improving hypernymy predictions by augmenting prompts with additional information through automatically identified co-hyponyms. An iterative approach is developed for predicting higher-level concepts, which further improves the quality on the BLESS dataset (MAP = 0.8). 
\end{abstract}

\maketitle

\section{Introduction}

Taxonomies are important tools for knowledge organization in various information systems and domains. A taxonomy describes the most significant relations between concepts: from more specific concepts (subclasses, hyponyms) to more general concepts (classes, hypernyms). Since it is quite difficult to create taxonomies from scratch, a large amount of research is devoted to extracting taxonomic relationships from text data (\cite{gom04,asw20}). Over the last 30 years, taxonomy construction techniques have been tested in several task settings, such as: evaluation of hypernym extraction on specially created datasets of positive (hypernym) and negative (other) relations (\cite{rol18,shw17}); hypernym discovery from a given text collection (\cite{cam18,ber18}); enrichment of an existing taxonomy (\cite{jurg16,niki20}); and taxonomy induction from scratch (\cite{yan09,snow06,bor15}).

Variety of methods proposed for hypernym acquisition tasks includes linear (so-called Hearst) patterns (\cite{hrst92,sei2016,rol18}) and syntactic patterns (\cite{snow04,ald18}), unsupervised \cite{{shw17}} and supervised vector-based techniques (\cite{rol14}) such as projection-learning (\cite{fu14,ber18,bai21}), etc. Currently, large language models based on neural transformer architectures (\cite{dev19,rad19,bro20}) allow studying novel types of methods for hypernym prediction and taxonomy construction based on special questions (prompts) to models \cite{sch21}. Prompts can be applied in zero-shot, fine-tuning and prompt tuning regimes (\cite{xu22}). In this paper, we experiment with zero-shot prompt techniques and study several types of prompt combinations on hypernym prediction datasets. We test the prompt combination techniques using several GPT-like models (\cite{rad19,bro20}). We investigate the following research questions:

RQ1: How consistent is the behavior of the language models on the set of prompts for predicting hypernyms compared to classical patterns?

RQ2: Can hypernym prediction using prompts benefit from using co-hyponym prompts?

RQ3: Is it possible to improve the quality of hypernym chain prediction by combining prompts or using them iteratively, going up step by step in the zero-shot regime?
\section{Related Works}
\subsection{Pattern-based approaches}
One of the most influential approaches to hypernym detection is the pattern-based approach. Its key idea is to exploit certain lexico-syntactic patterns to detect hypernym relations in text. For instance, patterns like “NPh such as NPt”, or “NPt and other NPh” often indicate hypernymy relations, where T is a hyponym and H is a hypernym, so-called Hearst patterns (\cite{hrst92}).  Such patterns may be created manually, or they may be learned automatically (\cite{snow04,shw16}). 

The lexico-syntactic patterns usually allow for a high precision of taxonomic relation extraction; however, this approach also has several problems that reduce the recall of relationship extraction. In the text collection used, there may not be a single explicit mention of contexts of the required type, despite the fact that X and Y are in the required relationship with each other (so-called sparsity of patterns \cite{rol18}).

To increase the recall of lexico-syntactic patterns, several methods have been proposed. In a number of works  an extended set of lexico-syntactic patterns is used. For example, the authors of \cite{sei2016}  use a set of 59 patterns collected from various sources. In this study, a large Internet corpus was processed, resulting in the extraction of more than 400 million taxonomic relationships. However, as the variety of patterns increases, the precision of extracted relationships decreases due to less definite patterns. \cite{sei2016}  found that some patterns are highly accurate. For example, such patterns as: “NPt and any other NPh”, “NPt and other NPh” have more than 0.7 precision. Some other patterns have very low precision, less than 0.2, for example,"such as "NPh like NPt".

 Another approach to improve the recall of patterns is proposed in \cite{rol18}. The authors apply 10 Hearst patterns to the combined Gigaword and Wikipedia text collection and create the matrix describing ppmi weights (positive pointwise mutual information) for words to be met in the patterns. This matrix is sparse, with a lot of zero values. Then they apply Singular Value Decomposition to reduce the matrix dimensionality. As a result, they achieve a significant improvement in hypernym detection.

Several papers discuss the use of co-hyponym patterns as an additional source of information for hypernym detection (\cite{wan19,ber18}). Co-hyponyms are words having the same hypernym. Co-hyponym patterns include so-called enumeration patterns as "X1 and X2", "X1, X2 and X3". \cite{wan19} supposed that if $x_i$ and $x_j$ match a “Such-As” or “Co-Hyponym” pattern, there is a large probability that no is-a relation exists between $x_i$ and $x_j$. \cite{ber18} extract co-hyponyms for a given query word, they suppose that hypernyms of co-hyponyms can be considered as candidate hypernyms for the initial query word.

\subsection{Unsupervised vector-based approaches}
The second main approach of hypernym extraction is based on the methods of distributional semantics and the so-called distributional hypothesis, which consists in the fact that semantic similarity between words (or other linguistic units) can be modeled based on the comparison (similarity) of the contexts of these words (\cite{len08}). It is assumed that the similarity of word contexts correlates with the semantic similarity of words. Hyponyms and hypernyms are semantically similar, therefore they should have similar context representations in the form of vectors.

In unsupervised approaches, it is assumed that there is some transformation of the original word vectors that better matches the hyponym-hypernym relationship. In particular, it can be assumed that hypernym-word contexts include most of the hyponym-word contexts—the so-called distributional inclusion hypothesis (\cite{rol14,len12}).

Another hypothesis that can be additionally utilized to predict hypernyms without training is the distributional exclusivity hypothesis, which estimates how much more diverse hypernym contexts are in comparison to hyponym contexts, for instance, in the invCL formula (\cite{len12}).

According to the distributional informativeness hypothesis, hypernyms are generally less informative than hyponyms, as they are likely to occur in more general contexts than their hyponyms, so called SLQS measure (\cite{san15}).

In the article \cite{shw17}, the authors investigated different unsupervised distributional measures and compared them ith supervised methods of vector-based hypernym prediction. They discovered that, in general, supervised methods surpass unsupervised ones; however, the former are susceptible to the distribution of training instances, which negatively impacts their reliability. Being based on general linguistic hypotheses and independent from training data, unsupervised measures are more robust.

\subsection{Zero-shot prompts for large language models}
Prompts for prediction hypernyms with large language models, such as BERT or GPT, can be based on classical lexico-syntactic patterns. For BERT, the probability of words in the hypernym position is estimated, and the most probable word is extracted (\cite{rav20, han21}). In GPT-like models, the probability of generating words in the hypernym position can be estimated, as well as the overall probability of the entire pattern can be calculated.

\cite{rav20} investigate whether BERT has a systematic understanding of hypernymy. They find that BERT fails when tasked with finding the hypernym of a hyponym in the plural; it also fails more often when the hyponym in question is uncommon, or has not been seen with its hypernym in Wikipedia. \cite{han21} study a prompting methodology to ask BERT what the hypernym of a given word is. They use prompts in the form of lexico-syntactic patterns, mask the position of the hypernym, and extract the most probable words predicted with the BERT model. The approach is tested on the Semeval-2018 hypernym discovery task. Several prompts were used. It was found that BERT with prompting outperforms other unsupervised models for hypernym discovery.

\cite{jai22} compares Masked Language Model (MLM) prompting (BERT, RoBERTa) and the overall estimation of whole prompt sentences in GPT-2 and masked models. They test prompt approaches in taxonomy induction SemEval tasks TExEval-1 (\cite{bor15}) and TExEval-2 (\cite{bor16}). They used the equipment taxonomy from TExEval-1 and the English-language environment, science and food taxonomies from TExEval-2. Analyzing the performance of different prompts, they found that prompts that unambiguously encode hypernymy are better: i.e., the type prompt is better when compared to noise-inducing prompts such as "is a" or "is kind of". Additionally, the most frequent prompts in pretraining corpora were the most competitive.



\cite{lia23} compares the possibility of large language models to predict hypernyms for abstract and physical nouns. To achieve this, they create a novel dataset with separate sets of words from two WordNet taxonomies: abstract concepts and physical concepts. Several models, including BERT, T5, ChatGPT, and others, were tested. All models showed worse results in predicting hypernyms for abstract nouns. The prompt to GPT was as follows: "Is {synset-1} a hypernym of {synset-2}? {synset1} means {the definition of synset-1}. {synset-2} means {the definition of synset-2}. Please directly answer YES or NO. (Do not return any explanation or any additional information.)". ChatGPT, with such a prompt, obtained 44.10 of F1 measure for abstract words and 73.04 for concrete words. This is much worse than the results achieved by other models, but they were used in a supervised setting.


\section{Approach}
We study an approach to exploiting prompts, which maps a pair of terms and a prompt type $p$ to a single sentence (\cite{jai22,ush22}):
\begin{equation}
    P_{kind}(alligator, reptile)="alligator \ is \ a \ kind \ of \ reptile"
\end{equation}

For a given term $t$ and a given prompt $p$, the goal is to retrieve its top $k$ most likely hypernyms from the list of candidates, estimating their probabilities using language models. Only GPT models are used in these experiments as they have previously been shown to fit better than BERT-like models (\cite{jai22}). Having a pair and some pattern, we compare estimations of the probability of either the entire sentence (\textit{full}, Formula (2)) or only for hypernym tokens (if they come last -- \textit{selective}, Formula (3)). We suppose that in some cases, an estimated probability of some prompt can be relatively low because, for example, it is not very frequent, but a hypernym is estimated as quite probable in the current prompt.


\begin{figure}
    \centering
    \includegraphics[scale=0.25]{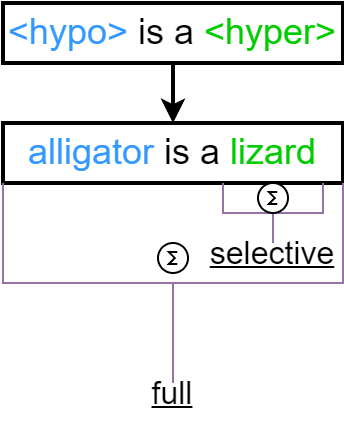}
    \vspace*{5mm}
    \caption{Probability calculation scheme for full and selective variants}
    \label{fig:enter-label}
\end{figure}

\begin{equation}
    score = exp(\sum^{|W|}_{i=1}log P(w_i|W_{<i})),
\end{equation}

\begin{equation}
    score = exp(\sum^{|W|}_{i=k}log P(w_i|W_{<i})),
\end{equation}

where $|W|$ is the length of a prompt-based sentence, word sequence located from $W_k$ to $W_{|W|}$ is a candidate hypernym.

In previous works, the use of prompts for zero-shot hypernym prediction has been studied. In our research, we conduct an extensive evaluation of hypernym prompts using a variety of GPT-like models. Using the results of prompt-based hypernym prediction experiments, we compare them with the results based on manual prompts described in (\cite{sei2016}).

The main idea of our study is experiment with prompts combinations, which includes the following directions:
\begin{enumerate}
    \item Combinations of  hypernym prompts,
    \item Combinations of hypernym  and co-hyponym prompts,
    \item Iterative application of hypernym prompts.
\end {enumerate}


\section{Datasets and Models}
We experiment with the following datasets from the hypernymysuite benchmark
\footnote{https://github.com/facebookresearch/hypernymysuite}(\cite{rol18}):
\begin{itemize}
 \item noun-noun subset of the BLESS dataset (\cite{bar11}), which contains
hypernymy annotations for 200 concrete, mostly
unambiguous nouns. Negative pairs contain a mixture of co-hyponymy, meronymy, and random pairs. This version contains 14,542 total pairs with 1,337
positive examples, 
\item LEDS (\cite{bar12}, which consists of 2,770 noun
pairs balanced between positive hypernymy examples, and randomly shuffled negative pairs,
\item EVAL (\cite{san15}), containing 7,378 pairs in a mixture of hypernymy, synonymy, antonymy, meronymy, and adjectival relations. EVAL is notable for its absence of random
pairs,
\item SHWARTZ (\cite{shw17}), which was collected from a mixture
of WordNet, DBPedia, and other resources. A 52,578 pair subset excluding
multiword expressions is extracted from the initial dataset,
\item WBLESS (\cite{wee14}), a 1,668 pair subset
of BLESS, with negative pairs being selected from
co-hyponymy, random, and hyponymy relations.
\end{itemize}

It is important to note that for all evaluations on hypernymysuite, we used code and data from the authors' github. Both the validation and test parts of BLESS dataset were used to calculate the MAP, since no separate tuning was made to the validation. 

We used the following GPT models:
\begin{itemize}
\item   GPT-2  -- 124M parameters (\cite{rad19}),
\item GPT-2 (Medium) -- 355M parameter version of GPT-2,
\item GPT-2 (Large) -- 774M parameter version of GPT-2,
\item GPT-2 (XL) -- 1.5B parameter version of GPT-2,
\item GPT-NEO-1.3b --  1.3B parameter-based transformer model designed using EleutherAI's replication of the GPT-3 architecture (\cite{bla22}),
\item llama-7b -- 7B parameter-based   auto-regressive language model, based on the transformer architecture (\cite{tou23}),
\item llama-2-7b  -- second version of the llama family of models, trained on an expanded dataset: 2T tokens, 4k context length (\cite{tou23b}).

\end{itemize}

We evaluate prompts and models in two different task settings of hypernym prediction. The first setting for evaluation, similar to \cite{rol18}, is to create a ranked list of pairs for the entire dataset, where correct hyponym-hypernym pairs should be located at the top of the list. This task is evaluated using the Average Precision measure (AP). The second variant of the evaluation is to create such a ranked list for each target word in a dataset, evaluate them using Average Precision measure, and then find the mean hypernym prediction value -- Mean Average Precision (MAP).

\begin{equation}
\begin{array}{c}
\hspace{10mm}AP_{i} = \frac{1}{M} \sum_{i}^{n} prec_{i} \times I[y_{i} = 1], \\ \ \\
\hspace{10mm}MAP = \frac{1}{N} \sum_{i=1}^{N} AP_{i},
\end{array}
\end{equation}
where $N$ and $M$ are the number of predicted and ground truth values, respectively, $prec_i$ is the fraction of ground truth values in the predictions from 1 to $i$, $y_i$ is the label of the $i$-th answer in the ranked list of predictions, and $I$ is the indicator function.

\begin{table}[ht!]
\centering

\caption{Datasets vs Model (Best for dataset based on hypernymysuite mean)}
\label{tab:1}       

\resizebox{0.9\textwidth}{!}{ 

\begin{tabular}{l|l|c|c|c|c|c|c|c}

\hline
&&\multicolumn{1}{c|}{Other}&\multicolumn{6}{c}{hypernymysuite} \\
\hline
&& MAP & \multicolumn{5}{c|}{Detection (AP)}&AP \\
\hline
model type & model & BLESS & BLESS & EVAL & LEDS & SHWARTZ & WBLESS &  mean \\
\hline
\multirow{4}{*}{\begin{tabular}{@{}l@{}}vector \\ (\cite{shw17})\end{tabular}} & Cosine & - & 0.12 & 0.29 & 0.71 & 0.31 & 0.53 & 0.39\\
&WeedsPrec & - &  0.19 & 0.39 & 0.87 & 0.43 & 0.68 & 0.51\\
&invCL & - &  0.18 & 0.37 & 0.89 & 0.38 & 0.66 & 0.49\\
&SLQS & - &  0.15 & 0.35 & 0.6 & 0.38 & 0.69 & 0.43\\
\hline
\multirow{4}{*}{\begin{tabular}{@{}l@{}}pattern \\ (\cite{rol18})\end{tabular}} & p(x, y) & - &  0.49 & 0.38 & 0.71 & 0.29 & 0.74 & 0,52\\
&ppmi(x, y) & - &  0.45 & 0.36 & 0.7 & 0.28 & 0.72 & 0,50\\
&sp(x, y) & - &  0.66 & 0.45 & 0.81 & 0.41 & 0.91 & 0,64\\
&spmi(x, y) & - &  \textbf{0.76} & \textbf{0.48} & \textbf{0.84} & \textbf{0.44} & \textbf{0.96} & \textbf{0,69}\\
\hline
\multirow{7}{*}{prompt full} &gpt2& 0.565 & 0.446 & 0.371 & 0.815 & 0.466 & 0.850 & 0,590\\
&gpt2-medium & 0.575 & 0.425 & \textbf{0.390} & 0.782 & \textbf{0.488} & 0.837 & 0,584\\
&gpt2-large & 0.596 & 0.473 & \textbf{0.390} & 0.806 & 0.463 & 0.860 &  0,598\\
&gpt2-xl & 0.600 & 0.489 & 0.360 & 0.812 & 0.480 & 0.868 & 0,602\\
&gpt-neo-1.3b & 0.639 & 0.407 & 0.353 & 0.845 & 0.466 & 0.853 & 0,585 \\
&llama-7b & 0.665 & 0.482 & 0.361 & 0.892 & 0.445 & 0.883 & 0,613 \\
&llama-2-7b & \textbf{0.701} & \textbf{0.577} & 0.374 & \textbf{0.910} & 0.452 & \textbf{0.915} & \textbf{0,646}\\
\hline
\multirow{7}{*}{prompt selective} & gpt2 & 0.561 & 0.472 & 0.410 & 0.916 & 0.511 & 0.877 & 0,637 \\
&gpt2-medium & 0.599 & 0.528 & 0.417 & 0.949 & 0.519 & 0.883 & 0,659 \\
&gpt2-large & 0.601 & 0.522 & \textbf{0.425} & 0.957 & 0.524 & 0.876 & 0,661\\
&gpt2-xl & 0.603 & 0.529 & 0.407 & 0.950 & 0.531 & 0.879 & 0,659 \\
&gpt-neo-1.3b & 0.638 & 0.562 & 0.419 & 0.959 & \textbf{0.544} & 0.904 & 0,678\\
&llama-7b & 0.664 & 0.595 & 0.420 & \textbf{0.968} & 0.484 & 0.914 & 0,676 \\
&llama-2-7b & \textbf{0.700} & \textbf{0.647} & \textbf{0.421} & \textbf{0.971} & 0.463 & \textbf{0.930} & \textbf{0,686}\\
\hline
\end{tabular}
}
\end{table}

\section{Single prompts experiments}
\subsection{Hypernym prompts}
In this study, we investigated 76 prompts for hypernymy prediction (\cite{jai22, sei2016}). It is important to note that many of these prompts are variations of the same semantic constructs; however, as the results demonstrate, even minor changes can significantly impact the final quality.

Table \ref{tab:1} presents the results of the best prompt for each model in comparison with previous works on unsupervised hypernym prediction:  vector-based  hypernym prediction (\cite{shw17}) and pattern-based hypernym prediction (\cite{rol18}). 

We can observe that the more parameters a model has, the better its results are, in practice, for all models and datasets. Models with a large number of parameters (llama-7b and llama-2-7b) achieve the best results on average. The \textit{selective} variant of the hypernym probability estimation is superior to the \textit{full} variant in terms of AP measure.

It can be seen that even a small GPT2 model obtains better results than vector-based techniques. Pattern-based results, which are based on a large corpus and SVD factorization of the PPMI matrix (\cite{rol18}), are difficult to achieve. Only the largest models, such as llama-7b and llama-2-7b, can achieve the same results of hypernym prediction on the considered datasets.

Table \ref{tab:2} (\textit{selective}) and Table \ref{tab:3} (\textit{full}) show  20 best prompts and their results obtained by the llama-2-7b model. It is evident that a single prompt (\textbf{hypo} or some other \textbf{hyper}) achieves the best results on all datasets. The performance of other prompts can be very diverse.

Seeing such different results, we decided to compare the performance obtained using prompts with the results of corresponding lexico-syntactic patterns described in (\cite{sei2016}). Tables \ref{tab:corr_selective} and \ref{tab:corr_full} (see Appendix) show the results of hypernym prediction for prompts (column "score"), other columns show prompt results for different models. In the last lines, Pearson and Spearman correlations for prompts with patterns results are shown. We see that the correlation between results is significant, the best prompts correspond to the best patterns. However, there is no correspondence between the size of models and achieved correlation values.

\begin{table}[ht!]
\caption{Datasets vs Prompt (selective, top 20)}
\label{tab:2}       
\resizebox{\textwidth}{!}{ 
\begin{tabular}{l|c|c|c|c|c|c|c}
\hline
&\multicolumn{1}{c|}{Other}&\multicolumn{6}{c}{hypernymysuite} \\
\hline
& MAP & \multicolumn{5}{c|}{Detection (AP)}& \\
\hline
prompt & BLESS & BLESS & EVAL & LEDS & SHWARTZ & WBLESS & mean \\
\hline
\begin{tabular}{@{}l@{}}\textbf{hypo} or some other \textbf{hyper}\end{tabular} & \textbf{0.700} & \textbf{0.647} & 0.421 & \textbf{0.971} & 0.463 & \textbf{0.930} & \textbf{0.686} \\
\begin{tabular}{@{}l@{}}\textbf{hypo} or any other \textbf{hyper}\end{tabular} & 0.669 & 0.600 & \textbf{0.424} & 0.968 & \textbf{0.503} & 0.919 & 0.683 \\
\begin{tabular}{@{}l@{}}\textbf{hypo} and any other \textbf{hyper}\end{tabular} & 0.672 & 0.601 & 0.408 & 0.961 & 0.497 & 0.921 & 0.678 \\
\begin{tabular}{@{}l@{}}\textbf{hypo} (and-or) (any-some) other \textbf{hyper}\end{tabular} & 0.668 & 0.603 & 0.414 & 0.949 & 0.475 & 0.910 & 0.670 \\
\begin{tabular}{@{}l@{}}\textbf{hypo} or other \textbf{hyper}\end{tabular} & 0.630 & 0.565 & 0.376 & 0.969 & 0.371 & 0.910 & 0.638 \\
\begin{tabular}{@{}l@{}}\textbf{hypo} and some other \textbf{hyper}\end{tabular} & 0.614 & 0.532 & 0.319 & 0.934 & 0.373 & 0.907 & 0.613 \\
\begin{tabular}{@{}l@{}}\textbf{hypo} is example of \textbf{hyper}\end{tabular} & 0.598 & 0.505 & 0.319 & 0.936 & 0.367 & 0.889 & 0.603 \\
\begin{tabular}{@{}l@{}}\textbf{hypo} which is a class of \textbf{hyper}\end{tabular} & 0.562 & 0.463 & 0.314 & 0.935 & 0.361 & 0.870 & 0.588 \\
\begin{tabular}{@{}l@{}}\textbf{hypo} is a type of \textbf{hyper}\end{tabular} & 0.495 & 0.406 & 0.349 & 0.937 & 0.419 & 0.810 & 0.584 \\
\begin{tabular}{@{}l@{}}\textbf{hypo} is a \textbf{hyper}\end{tabular} & 0.437 & 0.342 & 0.345 & 0.873 & 0.587 & 0.757 & 0.581 \\
\begin{tabular}{@{}l@{}}\textbf{hypo} and other \textbf{hyper}\end{tabular} & 0.572 & 0.485 & 0.279 & 0.936 & 0.301 & 0.901 & 0.580 \\
\begin{tabular}{@{}l@{}}\textbf{hypo} are examples of \textbf{hyper}\end{tabular} & 0.558 & 0.453 & 0.283 & 0.925 & 0.335 & 0.897 & 0.579 \\
\begin{tabular}{@{}l@{}}\textbf{hypo} which is a example of \textbf{hyper}\end{tabular} & 0.564 & 0.485 & 0.288 & 0.900 & 0.339 & 0.875 & 0.578 \\
\begin{tabular}{@{}l@{}}\textbf{hypo} is an \textbf{hyper} that\end{tabular} & 0.469 & 0.390 & 0.328 & 0.836 & 0.443 & 0.851 & 0.570 \\
\begin{tabular}{@{}l@{}}\textbf{hypo} like other \textbf{hyper}\end{tabular} & 0.510 & 0.399 & 0.281 & 0.898 & 0.306 & 0.831 & 0.543 \\
\begin{tabular}{@{}l@{}}\textbf{hypo} which is a kind of \textbf{hyper}\end{tabular} & 0.445 & 0.346 & 0.289 & 0.890 & 0.372 & 0.750 & 0.530 \\
\begin{tabular}{@{}l@{}}\textbf{hypo} sort of \textbf{hyper}\end{tabular} & 0.391 & 0.262 & 0.323 & 0.862 & 0.421 & 0.703 & 0.514 \\
\begin{tabular}{@{}l@{}}\textbf{hypo} which is a type of \textbf{hyper}\end{tabular} & 0.395 & 0.292 & 0.295 & 0.885 & 0.376 & 0.717 & 0.513 \\
\begin{tabular}{@{}l@{}}\textbf{hypo}, a kind of \textbf{hyper}\end{tabular} & 0.379 & 0.279 & 0.296 & 0.885 & 0.381 & 0.711 & 0.510 \\
\begin{tabular}{@{}l@{}}\textbf{hypo} which is a (type-example-class-kind) of \textbf{hyper}\end{tabular} & 0.411 & 0.303 & 0.297 & 0.890 & 0.340 & 0.711 & 0.508 \\
\hline
\end{tabular}
}
\end{table}

\begin{table}[ht!]
\caption{Datasets vs Prompt (full, top 20)}
\label{tab:3}       
\resizebox{\textwidth}{!}{ 
\begin{tabular}{l|c|c|c|c|c|c|c}
\hline
&\multicolumn{1}{c|}{Other}&\multicolumn{6}{c}{hypernymysuite} \\
\hline
& MAP & \multicolumn{5}{c|}{Detection (AP)}& \\
\hline
prompt & BLESS & BLESS & EVAL & LEDS & SHWARTZ & WBLESS & mean \\
\hline
\begin{tabular}{@{}l@{}}\textbf{hypo} is an \textbf{hyper} that\end{tabular} & 0.645 & 0.560 & \textbf{0.432} & 0.820 & \textbf{0.617} & 0.896 & \textbf{0.665} \\
\begin{tabular}{@{}l@{}}\textbf{hypo} or some other \textbf{hyper}\end{tabular} & \textbf{0.701} & \textbf{0.577} & 0.374 & \textbf{0.910} & 0.452 & \textbf{0.915} & 0.645 \\
\begin{tabular}{@{}l@{}}\textbf{hypo} or any other \textbf{hyper}\end{tabular} & 0.670 & 0.507 & 0.380 & 0.895 & 0.467 & 0.896 & 0.629 \\
\begin{tabular}{@{}l@{}}\textbf{hypo} and any other \textbf{hyper}\end{tabular} & 0.673 & 0.488 & 0.368 & 0.874 & 0.465 & 0.883 & 0.616 \\
\begin{tabular}{@{}l@{}}\textbf{hypo} or other \textbf{hyper}\end{tabular} & 0.630 & 0.492 & 0.351 & 0.897 & 0.404 & 0.886 & 0.606 \\
\begin{tabular}{@{}l@{}}like any \textbf{hyper}, \textbf{hypo}\end{tabular} & 0.552 & 0.408 & 0.368 & 0.843 & 0.522 & 0.794 & 0.587 \\
\begin{tabular}{@{}l@{}}\textbf{hypo} and some other \textbf{hyper}\end{tabular} & 0.615 & 0.458 & 0.298 & 0.853 & 0.407 & 0.879 & 0.579 \\
\begin{tabular}{@{}l@{}}\textbf{hypo} (and-or) (any-some) other \textbf{hyper}\end{tabular} & 0.667 & 0.432 & 0.332 & 0.788 & 0.459 & 0.849 & 0.572 \\
\begin{tabular}{@{}l@{}}\textbf{hypo} is a type of \textbf{hyper}\end{tabular} & 0.495 & 0.373 & 0.327 & 0.873 & 0.423 & 0.800 & 0.559 \\
\begin{tabular}{@{}l@{}}\textbf{hypo} and other \textbf{hyper}\end{tabular} & 0.572 & 0.416 & 0.279 & 0.850 & 0.369 & 0.870 & 0.557 \\
\begin{tabular}{@{}l@{}}such \textbf{hyper} as \textbf{hypo}\end{tabular} & 0.522 & 0.354 & 0.363 & 0.809 & 0.456 & 0.803 & 0.557 \\
\begin{tabular}{@{}l@{}}\textbf{hyper} other than \textbf{hypo}\end{tabular} & 0.520 & 0.358 & 0.397 & 0.826 & 0.406 & 0.790 & 0.556 \\
\begin{tabular}{@{}l@{}}\textbf{hypo} is example of \textbf{hyper}\end{tabular} & 0.599 & 0.429 & 0.290 & 0.819 & 0.387 & 0.847 & 0.555 \\
\begin{tabular}{@{}l@{}}\textbf{hyper} such as \textbf{hypo}\end{tabular} & 0.501 & 0.342 & 0.353 & 0.831 & 0.400 & 0.767 & 0.539 \\
\begin{tabular}{@{}l@{}}\textbf{hyper} e.g. \textbf{hypo}\end{tabular} & 0.529 & 0.359 & 0.334 & 0.824 & 0.355 & 0.807 & 0.536 \\
\begin{tabular}{@{}l@{}}\textbf{hypo} are examples of \textbf{hyper}\end{tabular} & 0.558 & 0.388 & 0.263 & 0.798 & 0.373 & 0.850 & 0.534 \\
\begin{tabular}{@{}l@{}}\textbf{hypo} is a \textbf{hyper}\end{tabular} & 0.436 & 0.297 & 0.304 & 0.790 & 0.530 & 0.731 & 0.530 \\
\begin{tabular}{@{}l@{}}unlike other \textbf{hyper}, \textbf{hypo}\end{tabular} & 0.516 & 0.353 & 0.301 & 0.821 & 0.388 & 0.789 & 0.530 \\
\begin{tabular}{@{}l@{}}examples of \textbf{hyper} are \textbf{hypo}\end{tabular} & 0.542 & 0.390 & 0.298 & 0.786 & 0.348 & 0.829 & 0.530 \\
\begin{tabular}{@{}l@{}}\textbf{hyper} which is similar to \textbf{hypo}\end{tabular} & 0.437 & 0.292 & 0.368 & 0.818 & 0.404 & 0.742 & 0.525 \\
\hline
\end{tabular}
}
\end{table}


\subsection{Co-hyponym prompts}
\begin{table}[ht]
\centering
\caption{Cohyponym prompts: selective}
\label{tab:4}       
\resizebox{0.6\textwidth}{!}{ 
\begin{tabular}{l|c|c}
\hline
prompt & AP & MAP \\
\hline
selective&& \\
\hline
\textbf{hypo} or \textbf{cohypo}  & 0.715 & 0.847 \\
\textbf{hypo} and \textbf{cohypo}  & 0.767  & 0.870 \\
\textbf{hypo}, \textbf{cohypo} and other & 0.661  & 0.790 \\
\textbf{hypo}, \textbf{cohypo} or other & 0.661  & 0.790 \\
such as \textbf{hypo} or \textbf{cohypo} & 0.804 & 0.901 \\
such as \textbf{hypo} and \textbf{cohypo} &  0.800  & 0.899 \\
\textbf{hypo}, \textbf{cohypo} and other of the same type & 0.661  & 0.790 \\
\textbf{hypo} and \textbf{cohypo} are the same type  & 0.767  & 0.871 \\
\textbf{hypo} or \textbf{cohypo} are the same type & 0.715  & 0.847 \\
such as \textbf{hypo} and \textbf{cohypo} of the same type  & 0.801 & 0.898 \\
such as \textbf{hypo}, \textbf{cohypo} and other of the same type & \textbf{0.824} & \textbf{0.919} \\
\hline
full&& \\
\hline
\textbf{hypo} or \textbf{cohypo} & 0.662 & 0.847 \\
\textbf{hypo} and \textbf{cohypo}  & 0.713  & 0.870 \\
\textbf{hypo}, \textbf{cohypo} and other& 0.745  & 0.900 \\
\textbf{hypo}, \textbf{cohypo} or other & 0.686  & 0.882 \\
such as \textbf{hypo} or \textbf{cohypo}  & 0.742  & 0.901 \\
such as \textbf{hypo} and \textbf{cohypo}  & 0.739  & 0.899 \\
\textbf{hypo}, \textbf{cohypo} and other of the same type  & 0.697  & 0.882 \\
\textbf{hypo} and \textbf{cohypo} are the same type  & 0.758 & 0.903 \\
\textbf{hypo} or \textbf{cohypo} are the same type & 0.674  & 0.862 \\
such as \textbf{hypo} and \textbf{cohypo} of the same type  & 0.731  & 0.866 \\
such as \textbf{hypo}, \textbf{cohypo} and other of the same type & \textbf{0.819}  & \textbf{0.930} \\
\end{tabular}
}
\end{table}

In our study we considered four types of co-hyponym prompts based on so-called enumeration patterns (\cite{wan19,ber18}):
\begin{itemize}
\item enumeration patterns without left and right contexts: "hypo or cohypo",  "hypo and cohypo",
\item enumeration patterns with right contexts: "hypo, cohypo and other"; "hypo, cohypo or other", "hypo, cohypo and other of the same type", "hypo and cohypo are the same type", "hypo or cohypo are the same type",
\item enumeration patterns with left contexts:  "such as hypo or cohypo", "such as hypo and cohypo",
\item enumeration patterns with left and right contexts: "such as hypo and cohypo of the same type"; "such as hypo, cohypo and other of the same type".
\end{itemize}

It should be noted that right contexts do not influence on calculating probability in the \textit{selective} variant.

We evaluated the quality of the co-hyponym prompts on the BLESS dataset. The evaluation results on 11 prompts and llama-2 model are shown in Table \ref{tab:4}. The primary metric for selecting the best approach and prompts is MAP, as our main focus is to determine the relative order of co-hyponyms for a target word, rather than the general ordering for a list of words.

We observe that the left context of enumeration pattern enhances the quality of co-hyponym prediction in both variants of probability estimation (\textit{full} and \textit{selective}), adding right contexts in the \textit{full} variant of probability estimation improves the results of prediction. The prompt "such as hypo, cohypo, and others of the same type" demonstrated the best quality for both the full and selective approaches.


\section{Combinations}
\subsection{Combinations of hypernyms prompts}
In our study, we investigated if combining different hypernym prompts could enhance hypernym prediction. The prompts were combined by averaging the weights of the prompts for each word pair. Several prompts were selected from the top, and the most successful prompt was taken as the basis, with the rest added to check if this would improve the final quality. The results can be seen in Tables \ref{tab:6} and \ref{tab:7}. Essentially, this approach does not improve the ranking quality for llama-2, only in one case did such a combination still have a positive effect: \textit{full}, prompts "hypo or some other hyper" + "hypo is an hyper that". On smaller models, such as gpt2-medium, the situation is similar: only the same pair of prompts was useful.

\begin{table}[ht]
\caption{Hypernymy prompts combination: selective}
\label{tab:6}       
\resizebox{\textwidth}{!}{ 
\begin{tabular}{l|c|c|c|c|c|c|c}
\hline
&\multicolumn{1}{c|}{Other}&\multicolumn{6}{c}{hypernymysuite} \\
\hline
& MAP & \multicolumn{5}{c|}{Detection (AP)}& \\
\hline
prompt & BLESS & BLESS & EVAL & LEDS & SHWARTZ & WBLESS & mean \\
\hline
\begin{tabular}{@{}l@{}}\textbf{hypo} or some other \textbf{hyper}\end{tabular} & \textbf{0.700} & \textbf{0.647} & 0.421 & \textbf{0.971} & 0.463 & 0.930 & \textbf{0.686} \\
\hline
\hline
\begin{tabular}{@{}l@{}}\textbf{hypo} or some other \textbf{hyper} + \\ \textbf{hypo} or any other \textbf{hyper}\end{tabular} & 0.689 & 0.627 & \textbf{0.425} & \textbf{0.971} & \textbf{0.487} & 0.926 & \textbf{0.687} \\
\cline{1-1}
\begin{tabular}{@{}l@{}}\textbf{hypo} or some other \textbf{hyper} + \\ \textbf{hypo} and any other \textbf{hyper}\end{tabular} & 0.696 & 0.634 & 0.419 & 0.969 & 0.485 & 0.928 & \textbf{0.687} \\
\cline{1-1}
\begin{tabular}{@{}l@{}}\textbf{hypo} or some other \textbf{hyper} + \\ \textbf{hypo} is an \textbf{hyper} that\end{tabular} & 0.697 & \textbf{0.646} & 0.395 & 0.946 & 0.483 & \textbf{0.940} & 0.682 \\
\cline{1-1}
\begin{tabular}{@{}l@{}}\textbf{hypo} or some other \textbf{hyper} + \\ \textbf{hypo} is a \textbf{hyper}\end{tabular} & 0.608 & 0.535 & 0.397 & 0.954 & 0.556 & 0.880 & 0.664 \\
\cline{1-1}
\begin{tabular}{@{}l@{}}\textbf{hypo} or some other \textbf{hyper} + \\ \textbf{hypo} is example of \textbf{hyper}\end{tabular} & 0.681 & 0.608 & 0.381 & 0.966 & 0.431 & 0.926 & 0.662 \\
\cline{1-1}
\begin{tabular}{@{}l@{}}\textbf{hypo} or some other \textbf{hyper} + \\ \textbf{hypo} which is a example of \textbf{hyper}\end{tabular} & 0.667 & 0.601 & 0.365 & 0.959 & 0.412 & 0.924 & 0.652 \\
\cline{1-1}
\begin{tabular}{@{}l@{}}\textbf{hypo} or some other \textbf{hyper} + \\ \textbf{hypo} which is a class of \textbf{hyper}\end{tabular} & 0.651 & 0.579 & 0.374 & 0.965 & 0.409 & 0.912 & 0.648 \\
\cline{1-1}
\begin{tabular}{@{}l@{}}\textbf{hypo} or some other \textbf{hyper} + \\ \textbf{hypo} are examples of \textbf{hyper}\end{tabular} & 0.657 & 0.580 & 0.359 & 0.963 & 0.405 & 0.931 & 0.648 \\
\cline{1-1}
\begin{tabular}{@{}l@{}}\textbf{hypo} or some other \textbf{hyper} + \\ \textbf{hypo} like other \textbf{hyper}\end{tabular} & 0.630 & 0.544 & 0.352 & 0.954 & 0.382 & 0.903 & 0.627 \\
\cline{1-1}
\begin{tabular}{@{}l@{}}\textbf{hypo} or some other \textbf{hyper} + \\ \textbf{hyper} e.g. \textbf{hypo}\end{tabular} & 0.635 & 0.562 & 0.354 & 0.939 & 0.323 & 0.893 & 0.614 \\
\cline{1-1}
\begin{tabular}{@{}l@{}}\textbf{hypo} or some other \textbf{hyper} + \\ \textbf{hyper} such as \textbf{hypo}\end{tabular} & 0.604 & 0.504 & 0.344 & 0.938 & 0.317 & 0.873 & 0.595 \\
\hline
\end{tabular}
}
\end{table}

\begin{table}[ht]
\caption{Hypernymy prompts combination: full}
\label{tab:7}       
\resizebox{\textwidth}{!}{ 
\begin{tabular}{l|c|c|c|c|c|c|c}
\hline
&\multicolumn{1}{c|}{Other}&\multicolumn{6}{c}{hypernymysuite} \\
\hline
& MAP & \multicolumn{5}{c|}{Detection (AP)}& \\
\hline
prompt & BLESS & BLESS & EVAL & LEDS & SHWARTZ & WBLESS & mean \\
\hline
\begin{tabular}{@{}l@{}}\textbf{hypo} is an \textbf{hyper} that\end{tabular} & 0.645 & 0.560 & 0.432 & 0.820 & \textbf{0.617} & 0.896 & 0.665 \\
\begin{tabular}{@{}l@{}}\textbf{hypo} or some other \textbf{hyper}\end{tabular} & 0.701 & 0.577 & 0.374 & \textbf{0.910} & 0.452 & 0.915 & 0.645 \\
\hline
\hline
\begin{tabular}{@{}l@{}}\textbf{hypo} or some other \textbf{hyper} + \\ \textbf{hypo} is an \textbf{hyper} that\end{tabular} & \textbf{0.772} & \textbf{0.670} & \textbf{0.435} & 0.887 & 0.544 & \textbf{0.936} & \textbf{0.694} \\
\cline{1-1}
\begin{tabular}{@{}l@{}}\textbf{hypo} or some other \textbf{hyper} + \\ \textbf{hypo} or any other \textbf{hyper}\end{tabular} & 0.690 & 0.547 & 0.380 & 0.905 & 0.461 & 0.908 & 0.640 \\
\cline{1-1}
\begin{tabular}{@{}l@{}}\textbf{hypo} or some other \textbf{hyper} + \\ \textbf{hypo} and any other \textbf{hyper}\end{tabular} & 0.697 & 0.546 & 0.377 & 0.899 & 0.461 & 0.906 & 0.638 \\
\cline{1-1}
\begin{tabular}{@{}l@{}}\textbf{hypo} or some other \textbf{hyper} + \\ \textbf{hyper} e.g. \textbf{hypo}\end{tabular} & 0.686 & 0.535 & 0.371 & 0.896 & 0.417 & 0.895 & 0.623 \\
\cline{1-1}
\begin{tabular}{@{}l@{}}\textbf{hypo} or some other \textbf{hyper} + \\ \textbf{hyper} such as \textbf{hypo}\end{tabular} & 0.656 & 0.514 & 0.380 & 0.897 & 0.437 & 0.874 & 0.620 \\
\cline{1-1}
\begin{tabular}{@{}l@{}}\textbf{hypo} or some other \textbf{hyper} + \\ \textbf{hypo} is example of \textbf{hyper}\end{tabular} & 0.682 & 0.536 & 0.338 & 0.886 & 0.426 & 0.902 & 0.618 \\
\cline{1-1}
\begin{tabular}{@{}l@{}}\textbf{hypo} or some other \textbf{hyper} + \\ \textbf{hypo} are examples of \textbf{hyper}\end{tabular} & 0.658 & 0.516 & 0.323 & 0.879 & 0.415 & 0.908 & 0.608 \\
\cline{1-1}
\begin{tabular}{@{}l@{}}\textbf{hypo} or some other \textbf{hyper} + \\ \textbf{hypo} is a \textbf{hyper}\end{tabular} & 0.607 & 0.464 & 0.350 & 0.875 & 0.494 & 0.856 & 0.608 \\
\cline{1-1}
\begin{tabular}{@{}l@{}}\textbf{hypo} or some other \textbf{hyper} + \\ \textbf{hypo} which is a class of \textbf{hyper}\end{tabular} & 0.652 & 0.504 & 0.340 & 0.868 & 0.414 & 0.879 & 0.601 \\
\cline{1-1}
\begin{tabular}{@{}l@{}}\textbf{hypo} or some other \textbf{hyper} + \\ \textbf{hypo} which is a example of \textbf{hyper}\end{tabular} & 0.669 & 0.495 & 0.330 & 0.844 & 0.406 & 0.889 & 0.593 \\
\cline{1-1}
\begin{tabular}{@{}l@{}}\textbf{hypo} or some other \textbf{hyper} + \\ \textbf{hypo} like other \textbf{hyper}\end{tabular} & 0.631 & 0.478 & 0.327 & 0.864 & 0.414 & 0.865 & 0.590 \\
\hline
\end{tabular}
}
\end{table}

\subsection{Co-hyponym-augmented prompts}
The concept of combining co-hyponyms with hypernyms prompts is to provide additional context to the prompt by listing its co-hyponyms along with the target word. The approach consists of two stages:
\begin{enumerate}
    \item Generating a list of co-hyponyms for target words,
    \item Modification of existing prompts using co-hyponyms.
\end{enumerate}

\subsubsection{Co-hyponym search}
Only in rare cases do we have a list of synonyms or co-hyponyms for the words we are interested in, so we needed to implement an automatic technique to find them. The implemented approach utilizes the capabilities of both statistical vector models and modern large language models and consists of the following steps:
\begin{enumerate}
    \item For each target word, we search for a list of the most similar words based on a vector model (fasttext),
    \item Next, we filter  too similar words in spelling using Levenshtein distance,
    \item Then, we require candidate words to appear in the WordNet 3.0, this step was used because some artifacts and forms of the word that remained after previous filtering stages could remain in the list,
    \item Finally, we rerank the words using the llama-2 language model and the best co-hyponym prompt to find co-hyponyms among co-hyponyms candidates.
\end{enumerate}

Although the list of co-hyponyms obtained in this way is not ideal, the approach usually provides a suitable co-hyponym at the first position. An example is presented in Table \ref{tab:cohypo_example}, for each step only the top 10 words are given.
\begin{table}[ht!]
\caption{Co-hyponym search example for word: jeweller}
\label{tab:cohypo_example}       
\centering
\resizebox{0.6\textwidth}{!}{ 
\begin{tabular}{c|c|c}
\hline
fasttext (step 1)& after filtering (step 2\&3) & after rerank (step 4) \\
\hline
jeweler & goldsmith & watchmaker\\
jewellers & silversmith & artisan\\
goldsmith & goldsmiths & optician\\
jewellery & watchmaker & blacksmith\\
jewelers & milliner & goldsmith\\
silversmith & shopkeeper & craftsman\\
Jeweller & glassworker & shoemaker\\
Jewellers & pawnbroker & silversmith\\
jewler & glazier & sculptor\\
goldsmiths & engraver & dressmaker\\
\end{tabular}
}

\end{table}

\subsubsection{Experiment details}
We selected top 2 prompts from the list of the best prompts (which differs) for each of the approaches (\textit{full} and \textit{selective}) and formed variants of prompts with co-hyponyms with them. We also noticed that co-hyponym-augmented prompts do not perform better unless the hypernym is modified to be plural, we did it with inflect package\footnote{https://github.com/jaraco/inflect}. 

\begin{table}[ht]
\caption{Co-hyponym-augmented prompts (selective): variations of "\textbf{hypo} or some other \textbf{hyper}" and "\textbf{hypo} is example of \textbf{hyper}"}
\label{tab:8}       
\resizebox{\textwidth}{!}{ 
\begin{tabular}{l|c|c|c|c|c|c|c}
\hline
&\multicolumn{1}{c|}{Other}&\multicolumn{6}{c}{hypernymysuite} \\
\hline
& MAP & \multicolumn{5}{c|}{Detection (AP)}& \\
\hline
prompt & BLESS & BLESS & EVAL & LEDS & SHWARTZ & WBLESS & mean \\
\hline
\begin{tabular}{@{}l@{}}\textbf{hypo} or some other \textbf{hyper}\end{tabular} & \textbf{0.700} & \textbf{0.647} & 0.421 & \textbf{0.971} & 0.463 & \textbf{0.930} & 0.686 \\
\hline
\textbf{hypo}, \textbf{cohypo} or any other \textbf{hyper} & 0.682 & 0.639 & 0.487 & 0.967 & 0.503 & 0.924 & \textbf{0.704} \\
\textbf{hypo} or \textbf{cohypo} or any other \textbf{hyper} & 0.689 & 0.643 & \textbf{0.499} & 0.964 & 0.486 & 0.923 & \textbf{0.703} \\
\textbf{hypo} and \textbf{cohypo} or some other \textbf{hyper} & 0.682 & 0.634 & 0.486 & 0.958 & 0.509 & 0.922 & 0.702 \\
\textbf{hypo} and \textbf{cohypo} or any other \textbf{hyper} & 0.666 & 0.622 & 0.477 & 0.958 & \textbf{0.533} & 0.915 & 0.701 \\
\textbf{hypo}, \textbf{cohypo} or some other \textbf{hyper} & 0.666 & 0.611 & 0.481 & 0.963 & 0.470 & 0.921 & 0.689 \\
\textbf{hypo} or \textbf{cohypo} or some other \textbf{hyper} & 0.652 & 0.593 & 0.486 & 0.957 & 0.467 & 0.914 & 0.683 \\
\textbf{hypo}, \textbf{cohypo} or any other an \textbf{hyper} & 0.652 & 0.563 & 0.441 & 0.917 & 0.502 & 0.912 & 0.667 \\
\textbf{hypo} and \textbf{cohypo} or any other an \textbf{hyper} & 0.635 & 0.558 & 0.426 & 0.911 & 0.519 & 0.905 & 0.664 \\
\textbf{hypo} or \textbf{cohypo} or any other an \textbf{hyper} & 0.640 & 0.554 & 0.443 & 0.913 & 0.489 & 0.909 & 0.662 \\
\textbf{hypo}, \textbf{cohypo} or some other an \textbf{hyper} & 0.608 & 0.506 & 0.426 & 0.900 & 0.488 & 0.896 & 0.643 \\
\textbf{hypo} and \textbf{cohypo} or some other an \textbf{hyper} & 0.597 & 0.501 & 0.416 & 0.895 & 0.501 & 0.888 & 0.640 \\
\textbf{hypo} or \textbf{cohypo} or some other an \textbf{hyper} & 0.582 & 0.475 & 0.427 & 0.888 & 0.482 & 0.882 & 0.631 \\
\hline
\hline
\begin{tabular}{@{}l@{}}\textbf{hypo} is example of \textbf{hyper}\end{tabular} & 0.598 & 0.505 & 0.319 & 0.936 & 0.367 & 0.889 & 0.603 \\
\hline
\textbf{hypo}, \textbf{cohypo} or other are examples of \textbf{hyper} & \textbf{0.728} & \textbf{0.683} & \textbf{0.484} & 0.961 & \textbf{0.529} & \textbf{0.938} & \textbf{0.719} \\
\textbf{hypo}, \textbf{cohypo} and other are examples of \textbf{hyper} & 0.719 & 0.670 & 0.462 & \textbf{0.964} & 0.520 & 0.934 & 0.710 \\
\textbf{hypo}, \textbf{cohypo} are examples of \textbf{hyper} & 0.712 & 0.659 & 0.461 & \textbf{0.964} & 0.506 & 0.925 & 0.703 \\
\textbf{hypo} or \textbf{cohypo} are examples of \textbf{hyper} & 0.703 & 0.655 & 0.463 & 0.958 & 0.494 & 0.923 & 0.699 \\
\textbf{hypo} and \textbf{cohypo} are examples of \textbf{hyper} & 0.695 & 0.641 & 0.452 & 0.957 & 0.480 & 0.916 & 0.689 \\
\hline
\end{tabular}
}
\end{table}

\begin{table}[ht]
\caption{Co-hyponym-augmented prompts (full): variations of "\textbf{hypo} is an \textbf{hyper} that" and "\textbf{hypo} or some other \textbf{hyper}"}
\label{tab:9}       
\resizebox{\textwidth}{!}{ 
\begin{tabular}{l|c|c|c|c|c|c|c}
\hline
&\multicolumn{1}{c|}{Other}&\multicolumn{6}{c}{hypernymysuite} \\
\hline
& MAP & \multicolumn{5}{c|}{Detection (AP)}& \\
\hline
prompt & BLESS & BLESS & EVAL & LEDS & SHWARTZ & WBLESS & mean \\
\hline
\begin{tabular}{@{}l@{}}\textbf{hypo} is an \textbf{hyper} that\end{tabular} & 0.645 & 0.560 & 0.432 & 0.820 & \textbf{0.617} & 0.896 & 0.665 \\
\hline
\textbf{hypo}, \textbf{cohypo} are an \textbf{hyper} that & \textbf{0.779} & \textbf{0.648} & 0.440 & 0.875 & 0.511 & \textbf{0.935} & \textbf{0.682} \\
\textbf{hypo}, \textbf{cohypo} are such \textbf{hyper} that & 0.751 & 0.623 & \textbf{0.444} & 0.892 & 0.497 & 0.930 & 0.677 \\
\textbf{hypo}, \textbf{cohypo} are \textbf{hyper} that & 0.719 & 0.587 & 0.442 & \textbf{0.896} & 0.524 & 0.913 & 0.673 \\
\textbf{hypo}, \textbf{cohypo} are the \textbf{hyper} that & 0.727 & 0.578 & 0.440 & 0.894 & 0.520 & 0.920 & 0.671 \\
\textbf{hypo}, \textbf{cohypo} are such an \textbf{hyper} that & 0.776 & 0.607 & 0.426 & 0.852 & 0.494 & 0.935 & 0.663 \\
\textbf{hypo} and \textbf{cohypo} are an \textbf{hyper} that & 0.755 & 0.602 & 0.436 & 0.850 & 0.497 & 0.919 & 0.661 \\
\textbf{hypo} and \textbf{cohypo} are \textbf{hyper} that & 0.711 & 0.562 & 0.436 & 0.883 & 0.505 & 0.906 & 0.658 \\
\textbf{hypo} or \textbf{cohypo} are an \textbf{hyper} that & 0.731 & 0.568 & 0.441 & 0.856 & 0.489 & 0.908 & 0.652 \\
\textbf{hypo} and \textbf{cohypo} are the \textbf{hyper} that & 0.706 & 0.546 & 0.431 & 0.881 & 0.496 & 0.902 & 0.651 \\
\textbf{hypo} or \textbf{cohypo} are such \textbf{hyper} that & 0.718 & 0.559 & 0.437 & 0.881 & 0.469 & 0.905 & 0.650 \\
\textbf{hypo}, \textbf{cohypo} are a \textbf{hyper} that & 0.687 & 0.538 & 0.430 & 0.882 & 0.498 & 0.899 & 0.649 \\
\textbf{hypo} and \textbf{cohypo} are such \textbf{hyper} that & 0.711 & 0.558 & 0.423 & 0.864 & 0.478 & 0.906 & 0.646 \\
\textbf{hypo} or \textbf{cohypo} are the \textbf{hyper} that & 0.674 & 0.506 & 0.433 & 0.884 & 0.486 & 0.888 & 0.639 \\
\textbf{hypo} or \textbf{cohypo} are \textbf{hyper} that & 0.667 & 0.497 & 0.428 & 0.876 & 0.490 & 0.882 & 0.635 \\
\textbf{hypo} or \textbf{cohypo} are such an \textbf{hyper} that & 0.736 & 0.524 & 0.425 & 0.826 & 0.473 & 0.907 & 0.631 \\
\textbf{hypo} and \textbf{cohypo} are a \textbf{hyper} that & 0.668 & 0.493 & 0.430 & 0.861 & 0.478 & 0.886 & 0.630 \\
\textbf{hypo} and \textbf{cohypo} are such an \textbf{hyper} that & 0.725 & 0.528 & 0.414 & 0.812 & 0.481 & 0.911 & 0.629 \\
\textbf{hypo} or \textbf{cohypo} are a \textbf{hyper} that & 0.640 & 0.454 & 0.428 & 0.865 & 0.471 & 0.864 & 0.616 \\
\hline
\hline
\begin{tabular}{@{}l@{}}\textbf{hypo} or some other \textbf{hyper}\end{tabular} & \textbf{0.701} & \textbf{0.577} & 0.374 & \textbf{0.910} & 0.452 & \textbf{0.915} & \textbf{0.645} \\
\hline
\textbf{hypo}, \textbf{cohypo} or any other \textbf{hyper} & 0.683 & 0.522 & \textbf{0.428} & 0.900 & 0.453 & 0.907 & 0.642 \\
\textbf{hypo}, \textbf{cohypo} or some other \textbf{hyper} & 0.666 & 0.523 & 0.425 & 0.901 & 0.444 & 0.905 & 0.640 \\
\textbf{hypo} or \textbf{cohypo} or any other \textbf{hyper} & 0.688 & 0.526 & 0.425 & 0.904 & 0.442 & 0.901 & 0.640 \\
\textbf{hypo} and \textbf{cohypo} or some other \textbf{hyper} & 0.684 & 0.486 & 0.412 & 0.886 & 0.451 & 0.887 & 0.625 \\
\textbf{hypo} and \textbf{cohypo} or any other \textbf{hyper} & 0.666 & 0.481 & 0.410 & 0.882 & \textbf{0.459} & 0.882 & 0.623 \\
\textbf{hypo} or \textbf{cohypo} or some other \textbf{hyper} & 0.651 & 0.481 & 0.413 & 0.896 & 0.433 & 0.887 & 0.622 \\
\textbf{hypo}, \textbf{cohypo} or any other an \textbf{hyper} & 0.655 & 0.455 & 0.389 & 0.828 & 0.436 & 0.882 & 0.598 \\
\textbf{hypo} or \textbf{cohypo} or any other an \textbf{hyper} & 0.643 & 0.441 & 0.381 & 0.832 & 0.429 & 0.869 & 0.590 \\
\textbf{hypo}, \textbf{cohypo} or some other an \textbf{hyper} & 0.610 & 0.434 & 0.379 & 0.825 & 0.435 & 0.866 & 0.588 \\
\textbf{hypo} and \textbf{cohypo} or any other an \textbf{hyper} & 0.638 & 0.425 & 0.370 & 0.814 & 0.436 & 0.855 & 0.580 \\
\textbf{hypo} or \textbf{cohypo} or some other an \textbf{hyper} & 0.585 & 0.376 & 0.367 & 0.816 & 0.426 & 0.842 & 0.565 \\
\textbf{hypo} and \textbf{cohypo} or some other an \textbf{hyper} & 0.599 & 0.383 & 0.360 & 0.805 & 0.433 & 0.832 & 0.562 \\
\hline
\end{tabular}
}
\end{table}

\subsubsection{Co-hyponym-augmented prompts results}
Results can be seen in Tables \ref{tab:8} and \ref{tab:9}. In the case of the \textit{full} probability calculation, it was possible to significantly improve the quality of the prompt "hypo is an hyper that", but there are no improvements for the prompt "hypo or some other hyper". For \textit{selective} variant there are also improvements for prompt "hypo is example of hyper" and  small positive changes in AP (mean column) for "hypo or some other hyper". Also, as can be see, variants with a comma between a word and its co-hyponym are on average superior to variants with “and” and “or”. In addition, it was unexpected that the presence of the article "an" for the prompt “hypo, cohypo are an hyper that” improves the quality.

\subsection{Iterative approach to ranking a list of hypernyms}
\subsubsection{Motivation}
Hypernyms can form transitive chains: if word $B$ is a hypernym of word $A$ and word $C$ is a hypernym of word $B$, than word $C$ is a hypernym of word $A$. In resources such as WordNet, hypernym chains can include 5 and more hypenyms for the target word. Hypernymy prompt for words that are distantly located in the hypernymy chain can obtain low probability estimation from language models because they are less frequent, have less observation in corpora. Therefore we study an iterative approach to hypernym prediction: obtaining most probable hypernym B for the target word A, we than apply hypernym prompts to this hypernym $B$ supposing that the most probable hypernym $C$ for hypernym $B$ is also a hypernym for the target word $A$. Similar idea in projection learning (search for recurrent hypernym projections) improved the results in hypernym discovery (\cite{bai21}).

\begin{table}[ht]
\centering
\caption{Iterations: superscript \textbf{+} when algorithm improves results, superscript \textbf{-} otherwise}
\label{tab:10}       
\begin{adjustbox}{max width=0.7\textwidth,max totalheight=\textheight,keepaspectratio}
\begin{tabular}{l|l|c|c|c}

\hline
&prompt & step 0 & step last & step mean \\
\hline
&selective&&& \\
\hline
\multirow{2}{*}{hyper}&\textbf{hypo} or some other \textbf{hyper} & 0.699 & 0.751\textsuperscript{+} & 0.757\textsuperscript{+} \\
&\textbf{hypo} is example of \textbf{hyper} & 0.597 & 0.612\textsuperscript{+} & 0.618\textsuperscript{+} \\
\cline{1-1}
\multirow{2}{*}{hyper + cohypo}&\textbf{hypo}, \textbf{cohypo} or any other \textbf{hyper} & 0.682 & 0.752\textsuperscript{+} & 0.750\textsuperscript{+} \\
&\textbf{hypo}, \textbf{cohypo} or other are examples of \textbf{hyper} & 0.729 & 0.773\textsuperscript{+} & \textbf{0.773}\textsuperscript{+} \\
\hline
&full&&& \\
\hline
\multirow{2}{*}{hyper}&\textbf{hypo} or some other \textbf{hyper} & 0.700 & 0.745\textsuperscript{+} & 0.760\textsuperscript{+} \\
&\textbf{hypo} is an \textbf{hyper} that & 0.646 & 0.588\textsuperscript{-} & 0.630\textsuperscript{-} \\
\cline{1-1}
\multirow{2}{*}{hyper + cohypo}&\textbf{hypo}, \textbf{cohypo} or any other \textbf{hyper} & 0.682 & 0.720\textsuperscript{+} & 0.733\textsuperscript{+} \\
&\textbf{hypo}, \textbf{cohypo} are an \textbf{hyper} that & 0.778 & 0.783\textsuperscript{+} & \textbf{0.801}\textsuperscript{+} \\
\hline
\end{tabular}
\end{adjustbox}
\end{table}

\subsubsection{Iterative approach: algorithm}

For the target word $w_0$ and selected prompt $pr$ the algorithm is as follows:
    
\begin{enumerate}
   \item The prompt $p$ is used to predict hypernyms for the target $w_0$ as described earlier,
   \item Algorithm steps are applied until a stopping criterion is reached. Step:
   \begin{enumerate}
       \item The list of candidates is ranked based on the mean score of past steps,
       \item The current target word $w_i$ is selected from the top of list of candidates that has not been used before,
       \item For the target word $w_i$ and the list of candidates, we calculate scores, but all previously selected words $[w_1, ..., w_{i-1}]$ receive a score of 0.0 (maximum),
       \item The stopping criterion is checked: the new maximum score is required to be higher than the maximum score at the previous step (except for the selected target word),
   \end{enumerate}
   \item The list is ranked based on the last score (column \textbf{step last}) or mean score (column \textbf{step mean}) among all steps that passed the criterion.
\end{enumerate}

\begin{figure}[ht!]
    \centering
    \includegraphics[scale=0.33]{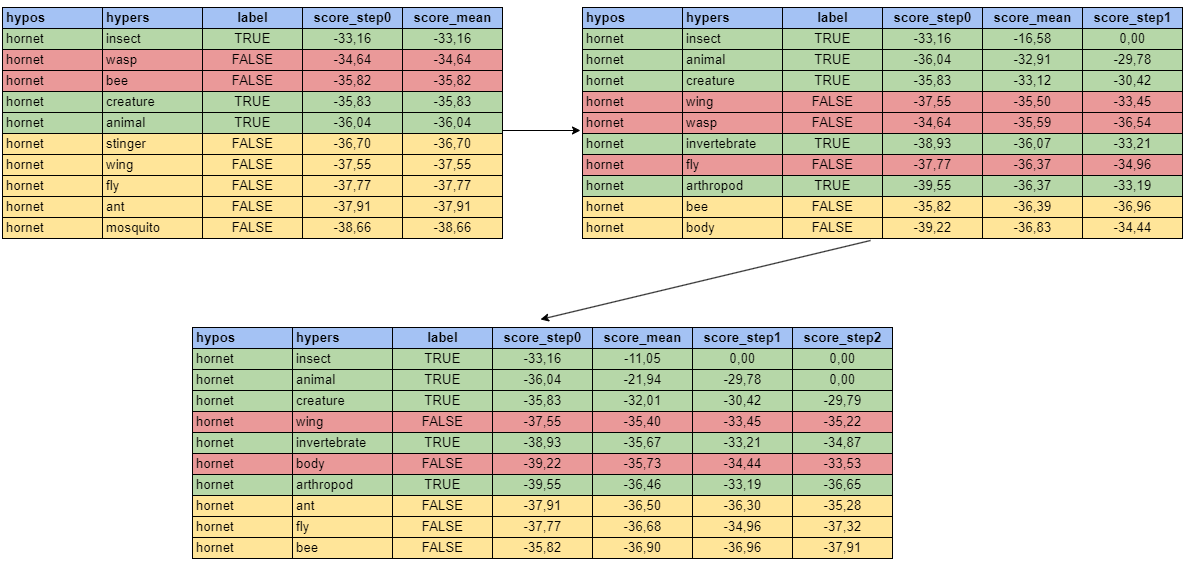}
    \vspace*{5mm}
    \caption{Example of iterative approach}
    \label{fig:iteration}
\end{figure}

The example in Figure \ref{fig:iteration} (only top-10 rows at each step) shows the iterative algorithm for the word \textbf{hornet}, which consists of 3 steps. In the first step, the target word $w_0$ is \textbf{hornet} and the entire list of candidates is evaluated by the llama-2 and prompt $p$. Column \textit{score\_mean} is calculated based on \textit{score\_step0} and $w\_1$ = \textbf{insect} is selected. Then on step 2 with $w\_1$ and other words from hypers column we calculate \textit{score\_step1}. Since the maximum value in \textit{score\_step1}  -29,78 is higher than -34,64 from \textit{score\_step0}, the criterion passes successfully and new \textit{score\_mean} are calculated. Based on \textit{score\_mean}, $w\_2$ = \textbf{animal} is selected. On step 3 we do the same as before and select $w\_3$ = \textbf{creature}, But after that, in step 4 the criterion fails (this step is not shown in the image) and the final ordering is formed based on the \textit{score\_mean} (average of \textit{score\_step0}, \textit{score\_step1}, \textit{score\_step2}).

\subsubsection{Iterative approach: results}
This approach is applicable only from the point of view of MAP metric, since the re-ranking is carried out separately for each target word. The results are presented in the Table \ref{tab:10}. The Table shows 2 basic prompts and their best variants with co-hyponyms for \textit{selective} and \textit{full} approaches. In most cases, the proposed iterative procedure improves the quality of ranking, often significantly. The best result 0.8 was achieved using the prompt "hypo, cohypo are an hyper that" with \textit{full} probability calculation approach. This is a significant increase in quality (from 0.7) compared to the simple prompts without co-hyponyms augmentation and the iterative procedure (see Table \ref{tab:2}, \ref{tab:3} and \ref{tab:10}).

\section{Conclusion}
This work explored a zero-shot approach to the hypernymy prediction task using large language models (LLMs). To do this, we used a method based on text probability calculation, applying it to various prompts. We did not evaluate instruct models (such as Vicuna, etc.), therefore the prompts were formed in accordance with the language, rather than chat or instruction format. We used two approaches to calculating the ranking score: \textit{full}, which uses the entire text sequence, and \textit{selective}, which uses predictions only for certain words. The \textit{selective} approach consistently showed higher quality on the hypernymysuite dataset, but the reason is not that it is better than \textit{full} (which is not even true at all), but because the evaluation method for these datasets is not entirely correct. In our opinion, instead of evaluating the whole bunch of words together, it is better to evaluate the ordering of each target word separately, for example, using the MAP metric. In general, we recommend using the probability estimate of the entire sequence.

In the article, we raised and answered 3 research questions:

RQ1: How consistent is the behavior of the language models on the set of prompts for predicting hypernyms compared to classical patterns? Experiments have shown a strong correlation between the effectiveness of language model prompts and classic patterns. At the same time, the level of correlation does not greatly depend on the size of the model, which means that preliminary selection of prompts can be carried out using smaller models and only then moving on to larger models. 

RQ2: Can hypernym prediction using prompts benefit from using co-hyponym prompts? The work also explored prompts for predicting co-hyponyms. The results showed that the llama-2 model copes with this task with high quality. We used this to improve the quality of hypernymy predictions by augmenting prompts with additional information through automatically identified co-hyponyms. This procedure significantly improved the quality of zero-shot predictions, although it is important to note that it had no effect on some prompts.

RQ3: Is it possible to improve the quality of hypernym chain prediction by combining prompts or using them iteratively, going up step by step in the zero-shot regime? In this case  it is necessary to predict not only a direct hypernym but also other higher-level concepts. The iterative approach was developed in which ranking are formed step-by-step, moving up the taxonomy. Using this approach made it possible to further improve the quality on the BLESS dataset (MAP metric). 

The best quality on the BLESS dataset (MAP 0.8 from the 0.7 with straightforward approach) was achieved by using the \textit{full} method, co-hyponym-augmented prompt "hypo, cohypo are an hyper that" and the iterative approach.

\section*{Acknowledgements}
The work of Natalia Loukachevitch (ideas of work with prompt combinations, writing  the related work section, proofreading) is supported by a grant for research centers in the field of artificial intelligence, provided by the Analytical Center for the Government of the Russian Federation in accordance with the subsidy agreement (agreement identifier 000000D730321P5Q0002) and the agreement with the Ivannikov Institute for System Programming of the Russian Academy of Sciences dated November 2, 2021 No. 70-2021-00142. The work of Mikhail Tikhomirov (implementation, experiments, writing a description of all experiments in the article, conclusion section) was supported by Non-commercial Foundation for Support of Science and Education "INTELLECT".

\label{lastpage}
\clearpage
\appendix
\section{Correlation with previous statistic results}\label{appendix}
\begin{table}[H]
\caption{Correlation: selective}
\centering
\label{tab:corr_selective}       
\begin{adjustbox}{max width=\textwidth,max totalheight=0.93\textheight,keepaspectratio}
\begin{tabular}{l|c|c|c|c|c|c|c|c|c|c|c|c|c|c|c}
\hline
pattern & score & \multicolumn{2}{c|}{gpt2} & \multicolumn{2}{c|}{gpt2-medium} & \multicolumn{2}{c|}{gpt2-large} & \multicolumn{2}{c|}{gpt2-xl}& \multicolumn{2}{c|}{gpt-neo-1.3B} & \multicolumn{2}{c}{llama-7B}& \multicolumn{2}{c}{llama-2-7B} \\
\hline
&&mean&MAP&mean&MAP&mean&MAP&mean&MAP&mean&MAP&mean&MAP&mean&MAP  \\
\hline
\textbf{hypo} and any other \textbf{hyper} & 0.76 & 0.629 & 0.554 & 0.650 & 0.583 & 0.658 & 0.601 & 0.644 & 0.568 & 0.654 & 0.593 & 0.672 & 0.658 & 0.678 & 0.672 \\
\textbf{hypo} or other \textbf{hyper} & 0.7 & 0.609 & 0.531 & 0.629 & 0.565 & 0.626 & 0.558 & 0.626 & 0.578 & 0.635 & 0.598 & 0.642 & 0.633 & 0.638 & 0.630 \\
\textbf{hypo} and other \textbf{hyper} & 0.7 & 0.552 & 0.451 & 0.561 & 0.479 & 0.575 & 0.502 & 0.564 & 0.490 & 0.569 & 0.487 & 0.584 & 0.577 & 0.580 & 0.572 \\
such \textbf{hyper} as \textbf{hypo} & 0.58 & 0.424 & 0.387 & 0.413 & 0.400 & 0.456 & 0.514 & 0.454 & 0.494 & 0.435 & 0.439 & 0.417 & 0.396 & 0.419 & 0.413 \\
\textbf{hyper} such as \textbf{hypo} & 0.58 & 0.412 & 0.331 & 0.410 & 0.302 & 0.421 & 0.311 & 0.420 & 0.333 & 0.410 & 0.315 & 0.429 & 0.357 & 0.420 & 0.353 \\
\textbf{hypo} are a \textbf{hyper} & 0.57 & 0.523 & 0.376 & 0.524 & 0.393 & 0.530 & 0.418 & 0.530 & 0.411 & 0.504 & 0.360 & 0.506 & 0.398 & 0.507 & 0.395 \\
\textbf{hypo} and some other \textbf{hyper} & 0.54 & 0.544 & 0.430 & 0.532 & 0.419 & 0.570 & 0.490 & 0.548 & 0.459 & 0.580 & 0.503 & 0.609 & 0.603 & 0.613 & 0.614 \\
\textbf{hypo} which is called \textbf{hyper} & 0.5 & 0.405 & 0.190 & 0.404 & 0.206 & 0.402 & 0.200 & 0.392 & 0.201 & 0.386 & 0.183 & 0.366 & 0.201 & 0.348 & 0.194 \\
\textbf{hypo}, kinds of \textbf{hyper} & 0.45 & 0.496 & 0.353 & 0.496 & 0.344 & 0.499 & 0.358 & 0.468 & 0.295 & 0.490 & 0.333 & 0.445 & 0.323 & 0.455 & 0.354 \\
examples of \textbf{hyper} are \textbf{hypo} & 0.45 & 0.332 & 0.265 & 0.312 & 0.208 & 0.307 & 0.196 & 0.319 & 0.232 & 0.345 & 0.263 & 0.436 & 0.421 & 0.445 & 0.421 \\
\textbf{hypo} is a \textbf{hyper} & 0.44 & 0.518 & 0.344 & 0.516 & 0.348 & 0.532 & 0.392 & 0.515 & 0.340 & 0.541 & 0.404 & 0.561 & 0.445 & 0.581 & 0.437 \\
\textbf{hyper} other than \textbf{hypo} & 0.44 & 0.382 & 0.281 & 0.393 & 0.280 & 0.398 & 0.281 & 0.384 & 0.271 & 0.384 & 0.266 & 0.399 & 0.313 & 0.415 & 0.330 \\
\textbf{hyper} including \textbf{hypo} & 0.44 & 0.369 & 0.289 & 0.361 & 0.274 & 0.379 & 0.282 & 0.371 & 0.280 & 0.371 & 0.284 & 0.380 & 0.320 & 0.379 & 0.323 \\
\textbf{hypo} were a \textbf{hyper} & 0.42 & 0.524 & 0.378 & 0.516 & 0.375 & 0.525 & 0.407 & 0.522 & 0.403 & 0.486 & 0.320 & 0.447 & 0.301 & 0.457 & 0.307 \\
\textbf{hypo} was a \textbf{hyper} & 0.39 & 0.516 & 0.348 & 0.526 & 0.382 & 0.532 & 0.391 & 0.529 & 0.390 & 0.506 & 0.353 & 0.497 & 0.361 & 0.498 & 0.359 \\
\textbf{hypo}, one of the \textbf{hyper} & 0.38 & 0.510 & 0.296 & 0.528 & 0.351 & 0.525 & 0.354 & 0.527 & 0.380 & 0.503 & 0.324 & 0.511 & 0.424 & 0.497 & 0.416 \\
\textbf{hypo} is example of \textbf{hyper} & 0.36 & 0.511 & 0.383 & 0.529 & 0.432 & 0.540 & 0.447 & 0.524 & 0.404 & 0.536 & 0.438 & 0.600 & 0.579 & 0.603 & 0.598 \\
\textbf{hypo}, forms of \textbf{hyper} & 0.33 & 0.516 & 0.368 & 0.520 & 0.379 & 0.528 & 0.415 & 0.483 & 0.331 & 0.496 & 0.336 & 0.488 & 0.374 & 0.504 & 0.420 \\
\textbf{hyper} e.g. \textbf{hypo} & 0.33 & 0.421 & 0.386 & 0.420 & 0.353 & 0.430 & 0.362 & 0.434 & 0.399 & 0.420 & 0.363 & 0.455 & 0.434 & 0.461 & 0.450 \\
examples of \textbf{hyper} is \textbf{hypo} & 0.33 & 0.308 & 0.185 & 0.297 & 0.155 & 0.306 & 0.181 & 0.299 & 0.170 & 0.314 & 0.181 & 0.371 & 0.383 & 0.349 & 0.280 \\
\textbf{hypo} or the many \textbf{hyper} & 0.31 & 0.477 & 0.330 & 0.510 & 0.361 & 0.505 & 0.374 & 0.504 & 0.366 & 0.484 & 0.325 & 0.449 & 0.278 & 0.466 & 0.329 \\
\textbf{hypo} like other \textbf{hyper} & 0.31 & 0.489 & 0.357 & 0.498 & 0.370 & 0.510 & 0.415 & 0.516 & 0.417 & 0.515 & 0.421 & 0.552 & 0.519 & 0.543 & 0.510 \\
\textbf{hyper} for example \textbf{hypo} & 0.31 & 0.401 & 0.425 & 0.374 & 0.370 & 0.388 & 0.429 & 0.402 & 0.460 & 0.449 & 0.512 & 0.430 & 0.478 & 0.441 & 0.472 \\
\textbf{hyper} which is similar to \textbf{hypo} & 0.29 & 0.367 & 0.250 & 0.364 & 0.207 & 0.363 & 0.192 & 0.371 & 0.218 & 0.362 & 0.218 & 0.387 & 0.250 & 0.395 & 0.263 \\
\textbf{hyper} i.e. \textbf{hypo} & 0.29 & 0.370 & 0.268 & 0.357 & 0.237 & 0.343 & 0.220 & 0.346 & 0.225 & 0.367 & 0.250 & 0.366 & 0.304 & 0.371 & 0.302 \\
\textbf{hyper} which are similar to \textbf{hypo} & 0.28 & 0.355 & 0.230 & 0.347 & 0.180 & 0.343 & 0.173 & 0.350 & 0.185 & 0.346 & 0.205 & 0.354 & 0.214 & 0.365 & 0.235 \\
\textbf{hyper} notably \textbf{hypo} & 0.28 & 0.355 & 0.357 & 0.324 & 0.293 & 0.350 & 0.341 & 0.319 & 0.256 & 0.361 & 0.391 & 0.377 & 0.369 & 0.362 & 0.354 \\
\textbf{hypo} which is named \textbf{hyper} & 0.26 & 0.422 & 0.248 & 0.423 & 0.242 & 0.428 & 0.250 & 0.408 & 0.229 & 0.391 & 0.194 & 0.381 & 0.227 & 0.369 & 0.222 \\
\textbf{hyper} principally \textbf{hypo} & 0.26 & 0.321 & 0.246 & 0.300 & 0.216 & 0.322 & 0.248 & 0.291 & 0.194 & 0.321 & 0.267 & 0.319 & 0.253 & 0.306 & 0.224 \\
\textbf{hyper} in particular \textbf{hypo} & 0.25 & 0.351 & 0.311 & 0.348 & 0.305 & 0.343 & 0.324 & 0.350 & 0.361 & 0.339 & 0.308 & 0.350 & 0.356 & 0.331 & 0.284 \\
\textbf{hyper} example of this is \textbf{hypo} & 0.25 & 0.367 & 0.340 & 0.377 & 0.371 & 0.393 & 0.385 & 0.385 & 0.350 & 0.383 & 0.404 & 0.389 & 0.399 & 0.385 & 0.380 \\
\textbf{hyper} among them \textbf{hypo} & 0.23 & 0.342 & 0.320 & 0.313 & 0.184 & 0.310 & 0.203 & 0.338 & 0.295 & 0.355 & 0.316 & 0.342 & 0.286 & 0.327 & 0.265 \\
\textbf{hyper} mainly \textbf{hypo} & 0.22 & 0.339 & 0.281 & 0.328 & 0.284 & 0.342 & 0.266 & 0.322 & 0.255 & 0.320 & 0.245 & 0.322 & 0.263 & 0.332 & 0.290 \\
\textbf{hyper} except \textbf{hypo} & 0.22 & 0.364 & 0.320 & 0.369 & 0.296 & 0.382 & 0.318 & 0.392 & 0.333 & 0.378 & 0.286 & 0.412 & 0.383 & 0.396 & 0.344 \\
\textbf{hypo} are examples of \textbf{hyper} & 0.2 & 0.524 & 0.433 & 0.551 & 0.483 & 0.561 & 0.489 & 0.573 & 0.525 & 0.541 & 0.461 & 0.593 & 0.571 & 0.579 & 0.558 \\
\textbf{hyper} particularly \textbf{hypo} & 0.19 & 0.338 & 0.287 & 0.335 & 0.292 & 0.354 & 0.321 & 0.327 & 0.262 & 0.354 & 0.323 & 0.354 & 0.310 & 0.351 & 0.306 \\
\textbf{hyper} especially \textbf{hypo} & 0.19 & 0.358 & 0.295 & 0.358 & 0.317 & 0.366 & 0.284 & 0.350 & 0.267 & 0.383 & 0.346 & 0.380 & 0.324 & 0.383 & 0.339 \\
\textbf{hypo}, a kind of \textbf{hyper} & 0.18 & 0.489 & 0.326 & 0.518 & 0.357 & 0.514 & 0.359 & 0.509 & 0.351 & 0.499 & 0.328 & 0.516 & 0.395 & 0.510 & 0.379 \\
\textbf{hypo}, a form of \textbf{hyper} & 0.18 & 0.505 & 0.363 & 0.533 & 0.397 & 0.543 & 0.441 & 0.522 & 0.398 & 0.485 & 0.326 & 0.514 & 0.404 & 0.503 & 0.376 \\
\textbf{hypo} which sound like \textbf{hyper} & 0.18 & 0.392 & 0.193 & 0.386 & 0.195 & 0.394 & 0.207 & 0.400 & 0.227 & 0.399 & 0.210 & 0.374 & 0.215 & 0.351 & 0.172 \\
\textbf{hypo} sort of \textbf{hyper} & 0.18 & 0.509 & 0.304 & 0.485 & 0.291 & 0.499 & 0.291 & 0.515 & 0.347 & 0.535 & 0.358 & 0.465 & 0.300 & 0.514 & 0.391 \\
\textbf{hyper} examples of this are \textbf{hypo} & 0.18 & 0.334 & 0.256 & 0.361 & 0.295 & 0.355 & 0.319 & 0.379 & 0.339 & 0.381 & 0.351 & 0.397 & 0.425 & 0.382 & 0.364 \\
\textbf{hypo} as \textbf{hyper} & 0.17 & 0.462 & 0.282 & 0.473 & 0.332 & 0.490 & 0.347 & 0.486 & 0.333 & 0.477 & 0.304 & 0.483 & 0.354 & 0.486 & 0.411 \\
\textbf{hyper} types \textbf{hypo} & 0.17 & 0.319 & 0.211 & 0.315 & 0.211 & 0.316 & 0.196 & 0.328 & 0.237 & 0.290 & 0.142 & 0.321 & 0.208 & 0.324 & 0.232 \\
\textbf{hyper} like \textbf{hypo} & 0.17 & 0.349 & 0.245 & 0.320 & 0.183 & 0.328 & 0.220 & 0.324 & 0.219 & 0.324 & 0.208 & 0.319 & 0.230 & 0.332 & 0.249 \\
\textbf{hyper} compared to \textbf{hypo} & 0.17 & 0.348 & 0.209 & 0.327 & 0.164 & 0.325 & 0.148 & 0.328 & 0.162 & 0.318 & 0.181 & 0.304 & 0.143 & 0.322 & 0.148 \\
\textbf{hyper} mostly \textbf{hypo} & 0.16 & 0.338 & 0.273 & 0.326 & 0.250 & 0.341 & 0.266 & 0.328 & 0.261 & 0.316 & 0.198 & 0.321 & 0.230 & 0.333 & 0.249 \\
compare \textbf{hypo} with \textbf{hyper} & 0.16 & 0.413 & 0.180 & 0.397 & 0.158 & 0.395 & 0.163 & 0.387 & 0.143 & 0.377 & 0.150 & 0.363 & 0.143 & 0.359 & 0.135 \\
\textbf{hypo}, one of those \textbf{hyper} & 0.15 & 0.503 & 0.360 & 0.511 & 0.385 & 0.524 & 0.408 & 0.525 & 0.399 & 0.496 & 0.343 & 0.471 & 0.364 & 0.466 & 0.355 \\
\textbf{hypo}, one of these \textbf{hyper} & 0.13 & 0.488 & 0.330 & 0.486 & 0.325 & 0.500 & 0.370 & 0.511 & 0.389 & 0.479 & 0.315 & 0.468 & 0.352 & 0.471 & 0.360 \\
\textbf{hypo} which look like \textbf{hyper} & 0.13 & 0.391 & 0.185 & 0.391 & 0.192 & 0.395 & 0.205 & 0.387 & 0.198 & 0.380 & 0.181 & 0.356 & 0.163 & 0.344 & 0.152 \\
\textbf{hyper} whether \textbf{hypo} or & 0.13 & 0.337 & 0.245 & 0.331 & 0.245 & 0.344 & 0.270 & 0.331 & 0.245 & 0.315 & 0.214 & 0.330 & 0.246 & 0.331 & 0.247 \\
\textbf{hyper} \textbf{hypo} for instance & 0.13 & 0.305 & 0.166 & 0.299 & 0.169 & 0.299 & 0.158 & 0.304 & 0.180 & 0.288 & 0.147 & 0.285 & 0.167 & 0.284 & 0.156 \\
\textbf{hypo} \textbf{hyper} types & 0.12 & 0.439 & 0.200 & 0.446 & 0.199 & 0.442 & 0.194 & 0.432 & 0.185 & 0.445 & 0.194 & 0.418 & 0.191 & 0.422 & 0.198 \\
\hline

Pearson correl & 1.0 & 0.523 & 0.625 & 0.496 & 0.576 & 0.509 & 0.568 & 0.491 & 0.549 & 0.541 & 0.596 & 0.574 & 0.601 & 0.569 & 0.611 \\
Spearman correl & 1.0 & 0.441 & 0.550 & 0.392 & 0.472 & 0.408 & 0.467 & 0.379 & 0.456 & 0.453 & 0.511 & 0.485 & 0.560 & 0.475 & 0.544 \\
\hline
\end{tabular}
\end{adjustbox}
\end{table}
\begin{table}[H]
\caption{Correlation: full}
\centering
\label{tab:corr_full}       
\begin{adjustbox}{max width=\textwidth,max totalheight=0.93\textheight,keepaspectratio}
\begin{tabular}{l|c|c|c|c|c|c|c|c|c|c|c|c|c|c|c}
\hline
pattern & score & \multicolumn{2}{c|}{gpt2} & \multicolumn{2}{c|}{gpt2-medium} & \multicolumn{2}{c|}{gpt2-large} & \multicolumn{2}{c|}{gpt2-xl}& \multicolumn{2}{c|}{gpt-neo-1.3B} & \multicolumn{2}{c}{llama-7B}& \multicolumn{2}{c}{llama-2-7B} \\
\hline
&&mean&MAP&mean&MAP&mean&MAP&mean&MAP&mean&MAP&mean&MAP&mean&MAP  \\
\hline
\textbf{hypo} and any other \textbf{hyper} & 0.76 & 0.547 & 0.554 & 0.557 & 0.584 & 0.579 & 0.601 & 0.578 & 0.569 & 0.555 & 0.593 & 0.595 & 0.659 & 0.616 & 0.673 \\
\textbf{hypo} or other \textbf{hyper} & 0.7 & 0.526 & 0.531 & 0.542 & 0.565 & 0.555 & 0.558 & 0.565 & 0.579 & 0.547 & 0.598 & 0.588 & 0.635 & 0.606 & 0.630 \\
\textbf{hypo} and other \textbf{hyper} & 0.7 & 0.482 & 0.452 & 0.489 & 0.479 & 0.518 & 0.502 & 0.514 & 0.490 & 0.501 & 0.488 & 0.546 & 0.579 & 0.557 & 0.572 \\
such \textbf{hyper} as \textbf{hypo} & 0.58 & 0.590 & 0.565 & 0.584 & 0.575 & 0.598 & 0.596 & 0.602 & 0.600 & 0.559 & 0.519 & 0.536 & 0.488 & 0.557 & 0.522 \\
\textbf{hyper} such as \textbf{hypo} & 0.58 & 0.490 & 0.391 & 0.531 & 0.433 & 0.525 & 0.456 & 0.532 & 0.479 & 0.511 & 0.388 & 0.571 & 0.576 & 0.539 & 0.501 \\
\textbf{hypo} are a \textbf{hyper} & 0.57 & 0.480 & 0.377 & 0.469 & 0.393 & 0.494 & 0.418 & 0.501 & 0.411 & 0.447 & 0.360 & 0.461 & 0.397 & 0.478 & 0.393 \\
\textbf{hypo} and some other \textbf{hyper} & 0.54 & 0.481 & 0.430 & 0.462 & 0.420 & 0.511 & 0.491 & 0.501 & 0.460 & 0.502 & 0.504 & 0.561 & 0.600 & 0.579 & 0.615 \\
\textbf{hypo} which is called \textbf{hyper} & 0.5 & 0.352 & 0.190 & 0.357 & 0.206 & 0.346 & 0.201 & 0.343 & 0.201 & 0.344 & 0.182 & 0.324 & 0.202 & 0.319 & 0.193 \\
\textbf{hypo}, kinds of \textbf{hyper} & 0.45 & 0.422 & 0.354 & 0.416 & 0.344 & 0.430 & 0.358 & 0.405 & 0.295 & 0.405 & 0.334 & 0.398 & 0.321 & 0.425 & 0.354 \\
examples of \textbf{hyper} are \textbf{hypo} & 0.45 & 0.417 & 0.322 & 0.390 & 0.261 & 0.386 & 0.267 & 0.403 & 0.292 & 0.410 & 0.314 & 0.507 & 0.505 & 0.530 & 0.542 \\
\textbf{hypo} is a \textbf{hyper} & 0.44 & 0.464 & 0.345 & 0.464 & 0.348 & 0.495 & 0.393 & 0.493 & 0.341 & 0.483 & 0.405 & 0.503 & 0.445 & 0.530 & 0.436 \\
\textbf{hyper} other than \textbf{hypo} & 0.44 & 0.484 & 0.402 & 0.501 & 0.414 & 0.498 & 0.412 & 0.456 & 0.352 & 0.499 & 0.381 & 0.545 & 0.520 & 0.556 & 0.520 \\
\textbf{hyper} including \textbf{hypo} & 0.44 & 0.448 & 0.339 & 0.455 & 0.387 & 0.473 & 0.382 & 0.462 & 0.411 & 0.465 & 0.354 & 0.485 & 0.479 & 0.468 & 0.421 \\
\textbf{hypo} were a \textbf{hyper} & 0.42 & 0.475 & 0.378 & 0.451 & 0.375 & 0.481 & 0.407 & 0.479 & 0.403 & 0.415 & 0.320 & 0.389 & 0.302 & 0.411 & 0.308 \\
\textbf{hypo} was a \textbf{hyper} & 0.39 & 0.454 & 0.349 & 0.458 & 0.383 & 0.485 & 0.392 & 0.489 & 0.390 & 0.432 & 0.353 & 0.421 & 0.360 & 0.449 & 0.360 \\
\textbf{hypo}, one of the \textbf{hyper} & 0.38 & 0.446 & 0.297 & 0.448 & 0.352 & 0.457 & 0.355 & 0.477 & 0.380 & 0.433 & 0.324 & 0.475 & 0.423 & 0.483 & 0.416 \\
\textbf{hypo} is example of \textbf{hyper} & 0.36 & 0.424 & 0.383 & 0.447 & 0.432 & 0.474 & 0.448 & 0.466 & 0.404 & 0.448 & 0.438 & 0.512 & 0.580 & 0.555 & 0.599 \\
\textbf{hypo}, forms of \textbf{hyper} & 0.33 & 0.431 & 0.368 & 0.449 & 0.379 & 0.469 & 0.415 & 0.438 & 0.332 & 0.422 & 0.336 & 0.425 & 0.372 & 0.457 & 0.419 \\
\textbf{hyper} e.g. \textbf{hypo} & 0.33 & 0.484 & 0.405 & 0.505 & 0.441 & 0.527 & 0.476 & 0.522 & 0.491 & 0.473 & 0.365 & 0.561 & 0.573 & 0.536 & 0.529 \\
examples of \textbf{hyper} is \textbf{hypo} & 0.33 & 0.394 & 0.293 & 0.365 & 0.237 & 0.367 & 0.250 & 0.368 & 0.265 & 0.373 & 0.251 & 0.437 & 0.408 & 0.433 & 0.367 \\
\textbf{hypo} or the many \textbf{hyper} & 0.31 & 0.423 & 0.330 & 0.434 & 0.362 & 0.431 & 0.375 & 0.437 & 0.366 & 0.408 & 0.326 & 0.394 & 0.279 & 0.428 & 0.330 \\
\textbf{hypo} like other \textbf{hyper} & 0.31 & 0.417 & 0.357 & 0.408 & 0.370 & 0.436 & 0.415 & 0.430 & 0.416 & 0.419 & 0.421 & 0.473 & 0.519 & 0.494 & 0.511 \\
\textbf{hyper} for example \textbf{hypo} & 0.31 & 0.434 & 0.328 & 0.417 & 0.289 & 0.426 & 0.346 & 0.429 & 0.362 & 0.470 & 0.380 & 0.520 & 0.519 & 0.510 & 0.509 \\
\textbf{hyper} which is similar to \textbf{hypo} & 0.29 & 0.427 & 0.260 & 0.441 & 0.278 & 0.451 & 0.293 & 0.454 & 0.303 & 0.442 & 0.269 & 0.519 & 0.408 & 0.525 & 0.437 \\
\textbf{hyper} i.e. \textbf{hypo} & 0.29 & 0.434 & 0.306 & 0.423 & 0.295 & 0.441 & 0.360 & 0.429 & 0.349 & 0.427 & 0.289 & 0.476 & 0.463 & 0.433 & 0.355 \\
\textbf{hyper} which are similar to \textbf{hypo} & 0.28 & 0.420 & 0.270 & 0.412 & 0.231 & 0.418 & 0.271 & 0.431 & 0.284 & 0.430 & 0.275 & 0.475 & 0.346 & 0.468 & 0.374 \\
\textbf{hyper} notably \textbf{hypo} & 0.28 & 0.407 & 0.304 & 0.385 & 0.275 & 0.402 & 0.310 & 0.375 & 0.299 & 0.421 & 0.326 & 0.460 & 0.399 & 0.451 & 0.384 \\
\textbf{hypo} which is named \textbf{hyper} & 0.26 & 0.361 & 0.249 & 0.371 & 0.243 & 0.365 & 0.250 & 0.355 & 0.228 & 0.349 & 0.194 & 0.332 & 0.226 & 0.332 & 0.222 \\
\textbf{hyper} principally \textbf{hypo} & 0.26 & 0.410 & 0.311 & 0.370 & 0.290 & 0.389 & 0.325 & 0.349 & 0.273 & 0.398 & 0.296 & 0.431 & 0.364 & 0.440 & 0.386 \\
\textbf{hyper} in particular \textbf{hypo} & 0.25 & 0.405 & 0.275 & 0.426 & 0.309 & 0.427 & 0.339 & 0.425 & 0.333 & 0.389 & 0.267 & 0.431 & 0.414 & 0.414 & 0.351 \\
\textbf{hyper} example of this is \textbf{hypo} & 0.25 & 0.397 & 0.287 & 0.405 & 0.265 & 0.438 & 0.342 & 0.431 & 0.335 & 0.445 & 0.325 & 0.457 & 0.412 & 0.453 & 0.381 \\
\textbf{hyper} among them \textbf{hypo} & 0.23 & 0.400 & 0.283 & 0.380 & 0.215 & 0.391 & 0.300 & 0.429 & 0.351 & 0.399 & 0.250 & 0.452 & 0.369 & 0.424 & 0.327 \\
\textbf{hyper} mainly \textbf{hypo} & 0.22 & 0.416 & 0.320 & 0.404 & 0.332 & 0.414 & 0.331 & 0.387 & 0.330 & 0.395 & 0.281 & 0.423 & 0.381 & 0.408 & 0.363 \\
\textbf{hyper} except \textbf{hypo} & 0.22 & 0.426 & 0.335 & 0.438 & 0.361 & 0.433 & 0.344 & 0.446 & 0.381 & 0.427 & 0.287 & 0.492 & 0.467 & 0.453 & 0.389 \\
\textbf{hypo} are examples of \textbf{hyper} & 0.2 & 0.444 & 0.433 & 0.457 & 0.483 & 0.493 & 0.489 & 0.501 & 0.526 & 0.454 & 0.462 & 0.504 & 0.570 & 0.534 & 0.558 \\
\textbf{hyper} particularly \textbf{hypo} & 0.19 & 0.408 & 0.298 & 0.415 & 0.343 & 0.410 & 0.332 & 0.377 & 0.311 & 0.403 & 0.268 & 0.443 & 0.428 & 0.439 & 0.377 \\
\textbf{hyper} especially \textbf{hypo} & 0.19 & 0.419 & 0.295 & 0.425 & 0.347 & 0.416 & 0.323 & 0.399 & 0.314 & 0.413 & 0.274 & 0.455 & 0.425 & 0.446 & 0.389 \\
\textbf{hypo}, a kind of \textbf{hyper} & 0.18 & 0.417 & 0.326 & 0.451 & 0.357 & 0.453 & 0.359 & 0.465 & 0.352 & 0.430 & 0.328 & 0.473 & 0.394 & 0.485 & 0.379 \\
\textbf{hypo}, a form of \textbf{hyper} & 0.18 & 0.442 & 0.363 & 0.478 & 0.397 & 0.495 & 0.441 & 0.489 & 0.399 & 0.435 & 0.327 & 0.485 & 0.404 & 0.490 & 0.377 \\
\textbf{hypo} which sound like \textbf{hyper} & 0.18 & 0.351 & 0.194 & 0.350 & 0.195 & 0.352 & 0.207 & 0.351 & 0.227 & 0.351 & 0.210 & 0.334 & 0.217 & 0.330 & 0.172 \\
\textbf{hypo} sort of \textbf{hyper} & 0.18 & 0.413 & 0.305 & 0.393 & 0.291 & 0.408 & 0.291 & 0.429 & 0.347 & 0.414 & 0.359 & 0.402 & 0.300 & 0.458 & 0.391 \\
\textbf{hyper} examples of this are \textbf{hypo} & 0.18 & 0.385 & 0.270 & 0.399 & 0.273 & 0.409 & 0.314 & 0.432 & 0.348 & 0.449 & 0.329 & 0.484 & 0.472 & 0.445 & 0.411 \\
\textbf{hypo} as \textbf{hyper} & 0.17 & 0.391 & 0.282 & 0.395 & 0.332 & 0.421 & 0.347 & 0.425 & 0.333 & 0.398 & 0.304 & 0.418 & 0.354 & 0.449 & 0.412 \\
\textbf{hyper} types \textbf{hypo} & 0.17 & 0.427 & 0.309 & 0.430 & 0.327 & 0.444 & 0.408 & 0.431 & 0.367 & 0.404 & 0.256 & 0.435 & 0.356 & 0.408 & 0.338 \\
\textbf{hyper} like \textbf{hypo} & 0.17 & 0.410 & 0.268 & 0.407 & 0.271 & 0.405 & 0.284 & 0.400 & 0.292 & 0.407 & 0.249 & 0.409 & 0.311 & 0.414 & 0.329 \\
\textbf{hyper} compared to \textbf{hypo} & 0.17 & 0.397 & 0.234 & 0.386 & 0.197 & 0.370 & 0.161 & 0.368 & 0.181 & 0.374 & 0.199 & 0.395 & 0.203 & 0.376 & 0.166 \\
\textbf{hyper} mostly \textbf{hypo} & 0.16 & 0.412 & 0.313 & 0.396 & 0.299 & 0.403 & 0.312 & 0.385 & 0.311 & 0.381 & 0.249 & 0.412 & 0.337 & 0.412 & 0.301 \\
compare \textbf{hypo} with \textbf{hyper} & 0.16 & 0.384 & 0.179 & 0.365 & 0.159 & 0.367 & 0.163 & 0.362 & 0.144 & 0.360 & 0.150 & 0.346 & 0.144 & 0.341 & 0.135 \\
\textbf{hypo}, one of those \textbf{hyper} & 0.15 & 0.442 & 0.360 & 0.423 & 0.385 & 0.450 & 0.407 & 0.463 & 0.399 & 0.415 & 0.343 & 0.420 & 0.364 & 0.439 & 0.356 \\
\textbf{hypo}, one of these \textbf{hyper} & 0.13 & 0.430 & 0.330 & 0.406 & 0.325 & 0.433 & 0.370 & 0.454 & 0.389 & 0.400 & 0.316 & 0.415 & 0.353 & 0.437 & 0.360 \\
\textbf{hypo} which look like \textbf{hyper} & 0.13 & 0.343 & 0.185 & 0.345 & 0.192 & 0.348 & 0.205 & 0.341 & 0.198 & 0.336 & 0.181 & 0.322 & 0.163 & 0.320 & 0.151 \\
\textbf{hyper} whether \textbf{hypo} or & 0.13 & 0.436 & 0.324 & 0.456 & 0.375 & 0.462 & 0.385 & 0.413 & 0.323 & 0.419 & 0.305 & 0.455 & 0.406 & 0.458 & 0.402 \\
\textbf{hyper} \textbf{hypo} for instance & 0.13 & 0.352 & 0.180 & 0.354 & 0.196 & 0.347 & 0.196 & 0.341 & 0.205 & 0.343 & 0.166 & 0.351 & 0.241 & 0.343 & 0.232 \\
\textbf{hypo} \textbf{hyper} types & 0.12 & 0.431 & 0.303 & 0.425 & 0.290 & 0.437 & 0.299 & 0.423 & 0.264 & 0.421 & 0.257 & 0.412 & 0.285 & 0.425 & 0.258 \\
\hline

Pearson correl & 1.0 & 0.700 & 0.689 & 0.660 & 0.629 & 0.654 & 0.617 & 0.642 & 0.603 & 0.704 & 0.708 & 0.593 & 0.585 & 0.631 & 0.621 \\
Spearman correl & 1.0 & 0.567 & 0.582 & 0.533 & 0.508 & 0.527 & 0.515 & 0.542 & 0.490 & 0.590 & 0.623 & 0.566 & 0.571 & 0.585 & 0.587 \\
\hline
\end{tabular}
\end{adjustbox}
\end{table}
\end{document}